\documentclass[conference]{IEEEtran}

\usepackage{cite}

\makeatletter
\def\endthebibliography{%
  \def\@noitemerr{\@latex@warning{Empty `thebibliography' environment}}%
  \endlist
}
\makeatother

\usepackage{graphics}
\usepackage{hyperref}
\usepackage{amsmath}
\usepackage{xspace}
\usepackage{booktabs}
\usepackage{subfigure}
\usepackage{url}
\usepackage[utf8]{inputenc}
\usepackage{newunicodechar}
\newunicodechar{⁻}{\textsuperscript{-}}
\usepackage{mathtools}
\usepackage{amsfonts}
\usepackage{amssymb}
\usepackage{multirow}
\usepackage{enumitem}
\usepackage{numprint}
\usepackage{tcolorbox}
\usepackage{xurl}
\usepackage{xstring}
\usepackage[T1]{fontenc}
\usepackage{graphicx}
\usepackage{setspace}
\usepackage{makecell}
\usepackage{xcolor,colortbl}

\usepackage{caption}
\usepackage{algorithm}
\usepackage{algpseudocode}

\usepackage{pifont}

\newcommand{\blackCircledNum}[1]{%
    \tikz[baseline=(char.base)]{
        \node[shape=circle, draw, fill=black, text=white, inner sep=0.5pt] (char) {\textbf{#1}};
    }%
}

\definecolor{gainsboro}{rgb}{0.86, 0.86, 0.86}
\definecolor{forest}{rgb}{0.86, 0.86, 0.86}
\definecolor{lightbrown}{rgb}{0.91, 0.94, 0.82} 
\definecolor{lightgray}{rgb}{0.93, 0.93, 0.93} 

\definecolor{CustomColor1}{RGB}{200, 217, 237} 
\definecolor{CustomColor2}{RGB}{196, 225, 165} 
\definecolor{CustomColor3}{RGB}{255, 213, 179} 

\definecolor{backgroundcolor}{RGB}{242, 248, 255}
\definecolor{rowgray}{gray}{0.95}
\definecolor{stagegray}{gray}{0.88}

\usepackage{tabularx}
\usepackage{ragged2e} 
\newcolumntype{Y}{>{\RaggedRight\arraybackslash}X} 
\usepackage{graphicx} 
\usepackage{hyperref}
\hypersetup{
    colorlinks=true,
    linkcolor=blue,
    filecolor=magenta,      
    urlcolor=blue,
    pdfpagemode=FullScreen,
    citecolor=blue,        
}

\usepackage{amsmath}

\usepackage{xcolor} 

\usepackage{rotating}

\usepackage{afterpage}
\usepackage{todonotes}
\setuptodonotes{inline, color=blue!30}

\usepackage[framemethod=TikZ]{mdframed}
\usepackage{algorithm}
\usepackage{algpseudocode}

\usepackage{booktabs}  
\usepackage{multirow}  

\usepackage{amssymb}  
\usepackage{pifont}   
\usepackage{fontawesome5}
\usepackage{wasysym}

\newcommand{\whitecircle}{\Circle}
\newcommand{\blackcircle}{\CIRCLE}  
\newcommand{\halfblackcircle}{\LEFTcircle}

\usepackage{svg}

\usepackage{amsthm}
\colorlet{shadecolor}{gray!15}
\colorlet{shadecolorr}{gray!10}
\usepackage[flushleft]{threeparttable}

\usepackage{booktabs}
\usepackage{multirow}

\definecolor{rowgray}{gray}{0.95}

\usepackage{amssymb} 

\newcommand{\iconcrypto}{\scalebox{1.3}{\ding{57}}}  
\newcommand{\iconecon}{\scalebox{1.3}{\ding{74}}}     
\newcommand{\icondetect}{\scalebox{1.3}{\ding{39}}}  
\newcommand{\icongov}{\scalebox{1.3}{\ding{45}}}     
\newcommand{\iconinfra}{\scalebox{1.3}{\ding{105}}}    

\newcounter{insight}
\renewcommand{\theinsight}{\arabic{insight}}

\newcommand{\insightbox}[1]{%
  \par\addvspace{0.5em}%
  \refstepcounter{insight}%
  \noindent
  \begingroup
  \setlength{\fboxsep}{6pt}%
  \setlength{\parindent}{0pt}%
  \setlength{\parskip}{0pt}%
  \fcolorbox{black}{shadecolorr}{%
    \begin{minipage}{0.94\columnwidth}%
      \textbf{Insight~\theinsight.}\hspace{0.6em}%
      \itshape #1%
    \end{minipage}%
  }%
  \endgroup
  \par\addvspace{0.5em}%
}

\newtheorem{gaps}{Gap}

\newcommand{\gapbox}[1]{%
  \par\addvspace{0.2em}%
  \noindent
  \begin{tikzpicture}
    \node[draw=black, dashed, fill=shadecolor, inner sep=6pt, rounded corners] (box) 
    {%
      \begin{minipage}{0.94\columnwidth}%
        \setlength{\parindent}{0pt}%
        \setlength{\parskip}{0pt}%
        \begin{gaps}\leavevmode
          #1%
        \end{gaps}%
      \end{minipage}%
    };
  \end{tikzpicture}%
}

\makeatletter
\newcommand{\ssymbol}[1]{^{\@fnsymbol{#1}}}
\makeatother

\usepackage[noextra]{acro}
\acsetup{single}

\DeclareAcronym{ZKP}{
  short = ZKP,
  long  = zero-knowledge proof,
}
\newcommand{\ZKP}{\ac{ZKP}\xspace}
\newcommand{\ZKPs}{\acp{ZKP}\xspace}

\DeclareAcronym{TEE}{
  short = TEE,
  long  = Trusted Execution Environment,
}

\newcommand{\TEEs}{\acp{TEE}\xspace}

\DeclareAcronym{FL}{
  short = FL,
  long  = Federated Learning,
}
\newcommand{\FL}{\ac{FL}\xspace}

\DeclareAcronym{AI}{
  short = AI,
  long  = Artificial Intelligence,
}
\newcommand{\AI}{\ac{AI}\xspace}

\DeclareAcronym{ML}{
  short = ML,
  long  = Machine Learning,
}
\newcommand{\ML}{\ac{ML}\xspace}

\DeclareAcronym{zkML}{
  short = zkML,
  long  = Zero Knowledge-Machine Learning,
}
\newcommand{\zkML}{\ac{zkML}\xspace}

\DeclareAcronym{opML}{
  short = opML,
  long  = Optimistic-Machine Learning,
}
\newcommand{\opML}{\ac{opML}\xspace}

\DeclareAcronym{DeFi}{
  short = DeFi,
  long  = Decentralized Finance,
}
\newcommand{\DeFi}{\ac{DeFi}\xspace}

\DeclareAcronym{DeAI}{
  short = DeAI,
  long  = Decentralized Artificial Intelligence,
  short-indefinite = (DeAI),
}
\newcommand{\DeAI}{\ac{DeAI}\xspace}

\DeclareAcronym{CeAI}{
  short = CeAI,
  long  = Centralized Artificial Intelligence,
}
\newcommand{\CeAI}{\ac{CeAI}\xspace}

\DeclareAcronym{PoW}{
  short = PoW,
  long  = Proof-of-Work,
}
\newcommand{\PoW}{\ac{PoW}\xspace}

\DeclareAcronym{PoS}{
  short = PoS,
  long  = Proof-of-Stake,
}
\newcommand{\PoS}{\ac{PoS}\xspace}

\DeclareAcronym{PoL}{
  short = PoL,
  long  = Proof-of-Learning,
}
\newcommand{\PoL}{\ac{PoL}\xspace}

\DeclareAcronym{uPoW}{
  short = uPoW,
  long  = Proof-of-Useful-Work,
}
\newcommand{\uPoW}{\ac{uPoW}\xspace}

\DeclareAcronym{LLM}{
  short = LLM,
  long  = Large Language Model,
}

\newcommand{\LLMs}{\acp{LLM}\xspace}

\DeclareAcronym{PtoP}{
  short = P2P,
  long  = Peer-to-Peer,
}
\newcommand{\PtoP}{\ac{PtoP}\xspace}

\DeclareAcronym{DAO}{
  short = DAO,
  long  = Decentralized Autonomous Organization,
}
\newcommand{\DAO}{\ac{DAO}\xspace}

\DeclareAcronym{SoK}{
  short = SoK,
  long = Systematization of Knowledge,
  short-indefinite = (SoK),
}
\newcommand{\SoK}{\ac{SoK}\xspace}

\DeclareAcronym{DML}{
  short = DML,
  long = Distributed Machine Learning,
}
\newcommand{\DML}{\ac{DML}\xspace}

\DeclareAcronym{DFL}{
  short = DFL,
  long = Decentralized Federated Learning,
}
\newcommand{\DFL}{\ac{DFL}\xspace}

\usepackage{graphicx}

\acsetup{
  first-style = long-short
}

\title {SoK: Blockchain-Based Decentralized AI (DeAI)}

\author{
\IEEEauthorblockN{
Elizabeth Lui\textsuperscript{*}\IEEEauthorrefmark{4},
Rui Sun\textsuperscript{*}\IEEEauthorrefmark{3}\IEEEauthorrefmark{6},
Vatsal Shah\IEEEauthorrefmark{4},
Xihan Xiong\IEEEauthorrefmark{2}, Jiahao Sun\IEEEauthorrefmark{4},\\
Davide Crapis\IEEEauthorrefmark{5},
William Knottenbelt\IEEEauthorrefmark{2},
Zhipeng Wang\textsuperscript{\scriptsize \protect\faEnvelope} \IEEEauthorrefmark{6}
}

\IEEEauthorblockA{
 \IEEEauthorrefmark{4}\textit{FLock.io}, 
 \IEEEauthorrefmark{6}\textit{The University of Manchester}\\
\IEEEauthorrefmark{3}\textit{Newcastle University}, 
\IEEEauthorrefmark{2}\textit{Imperial College London},
\IEEEauthorrefmark{5}\textit{Ethereum Foundation}
}
}

\begin{document}
\maketitle

\begin{abstract}
Centralization enhances the efficiency of \AI but also introduces critical challenges, such as single points of failure, inherent biases, data privacy risks, and scalability limitations. To address these issues, blockchain-based \Acl{DeAI} (\acs{DeAI}) has emerged as a promising paradigm to improve the trustworthiness of \AI systems. Despite rapid adoption in industry, the academic community lacks a systematic analysis of DeAI’s technical foundations, opportunities, and challenges. This work presents the first \Acl{SoK} (\acs{SoK}) on \DeAI, offering a formal definition, a taxonomy of existing solutions based on the \AI lifecycle, and an in-depth investigation of the roles of blockchain in enabling secure and incentive-compatible collaboration. We further review security risks across the \DeAI lifecycle and empirically evaluate representative mitigation techniques. Finally, we highlight open research challenges and future directions for advancing \DeAI in blockchain-based systems.

\end{abstract}

\begingroup
\renewcommand\thefootnote{*}
\footnotetext{These authors contributed equally to this work.}

\renewcommand\thefootnote{\protect\faIcon{envelope}}
\footnotetext{Corresponding author.}
\endgroup

\section{Introduction}
\Acl{CeAI} (\acs{CeAI}) systems underpin advances in vision~\cite{Krizhevsky2012ImageNet}, language~\cite{Devlin2019BERT}, and healthcare~\cite{esteva2019guide}. Despite these successes, centralization places data, compute, and training under the control of a single entity. This creates single points of failure~\cite{Brundage2018Malicious}, limits model diversity~\cite{challen2019artificial,bender2021Parrots}, raises privacy risks~\cite{Carlini2021Extracting,yao2024survey,das2024security}, and leads to scalability bottlenecks~\cite{Dean2012LargeScale} as well as constrained innovation~\cite{AImodelsDemocratization,montes2019distributed,guan2020artificial}. 

In this context, \DeAI has emerged as a promising alternative, often leveraging blockchain to coordinate \AI tasks without a central authority. Decentralization, transparency, and immutability, along with consensus protocols and smart contracts~\cite{bano2019sok,werner2022sok}, enable trust-minimized and verifiable data, compute, and training workflows. Native cryptocurrencies introduce incentive mechanisms to reward honest contributions, while on-chain verification can enforce provenance and accountability~\cite{Xie2024BlockchainIoT,Fernandez2018BlockchainIoT}, mitigating opacity and incentive misalignment in \CeAI. Although  \DeAI systems are widely adopted in industry~\cite{BittensorWhitepaper,flock-white-paper,vana-docs,sentient-docs,fetch-ai-white-paper,akash-white-paper,morpheus-white-paper}, integrating blockchain with \AI introduces challenges: \DeAI must confront scalability limits~\cite{Zheng2018BlockchainChallenges,Chen2020Scalable} and balance transparency with data privacy~\cite{Shrestha2020Privacy}. Resolving these tensions is essential for practical deployment. 

Despite rapid industrial growth, the academic community lacks a systematic analysis of \DeAI solutions. In response, this work provides an \SoK of blockchain-based \DeAI, systematically examining its architectures, challenges, and opportunities. We aim to address the following research questions:


\begin{itemize} [leftmargin = *]
\item \textbf{RQ1:} What is the formal definition and general taxonomy of blockchain-based \DeAI?

\item \textbf{RQ2:} How can blockchain be used to decentralize and secure \AI systems? 

\item \textbf{RQ3:} What insights and research gaps can be drawn from existing \DeAI solutions?

\item \textbf{RQ4:} What are the key security risks across the \DeAI lifecycle, and how effective are representative mitigations when evaluated empirically?

\end{itemize}

To the best of our knowledge, this is the first comprehensive systematization of blockchain-based \DeAI and its design trade-offs. We summarize our contributions as follows: 

\begin{itemize}[leftmargin = *]

    \item We provide a formal definition of blockchain-based \DeAI, including an identification of its essential properties.
    \item We introduce a taxonomy that summarizes existing \DeAI solutions and categorizes them based on the lifecycle of an \AI model, along with an analysis of their structural similarities and differences (see Table~\ref{tab:deai_blockchain}).

    \item We investigate the functionalities of blockchain in existing \DeAI solutions, analyze how blockchain features contribute to security, transparency, and trustworthiness, and enable fair incentives for contributors in decentralized systems.

    \item We synthesize key insights and identify research gaps in building blockchain-based \DeAI solutions, outlining directions for future work. In addition to the gaps discussed throughout the paper, we further supplement these with a set of open research questions provided in Appendix~\ref{app:open-research-challenges}.

    \item We present a comprehensive review of security risks in blockchain-based \DeAI, complemented by an empirical evaluation of representative mitigation techniques across the different stages of the \DeAI lifecycle.
\end{itemize}

\section{Background}

\subsection{Centralized, Distributed, and Federated Learning}
\AI systems aim to perform tasks that typically require human intelligence, such as learning, reasoning, problem-solving, and language understanding~\cite{russell2010ai, Nilsson2009Quest, Mitchell1997MachineLearning, LeCun2015DeepLearning, Krizhevsky2012ImageNet, Devlin2019BERT, Silver2016Mastering}. Most current \AI systems, however, remain centralized: model creation, training, and deployment are controlled by a single or a few entities (see Figure~\ref{fig:learning_para_stru}(B) in Appendix~\ref{app:additional-figures}).  While effective for building powerful applications, this centralized paradigm introduces challenges including computing bottlenecks, data availability, and privacy risks~\cite{Yang2019Federated}, value bias~\cite{Buolamwini2018Gender}, governance issues~\cite{chandrasekaran2021sok}, and broader ethical concerns~\cite{Jobin2019GlobalAI}.

Attempts have emerged to distribute data and training, albeit still within centralized systems. For instance, \textbf{\Acl{DML} (\acs{DML})} addresses the computational limits of centralized \Acl{ML} (\acs{ML}) with large models and datasets by distributing computation across multiple units for parallel execution~\cite{verbraeken2020survey}. \DML is typically categorized into \emph{data parallelism} and \emph{model parallelism}. In data parallelism (Figure~\ref{fig:learning_para_stru}(C)), the full model is replicated on each unit, which processes a data shard before aggregating results into a global update. In model parallelism, the model is partitioned across machines, enabling training of models that do not fit on a single device, but requiring all machines to access the full dataset, potentially increasing privacy risks.


\textbf{\Acl{FL} (\acs{FL})} provides a way to train large-scale models while preserving privacy. A naïve privacy solution is standalone on-device learning (Figure~\ref{fig:learning_para_stru}(A)), but this creates data silos that degrade model quality. FL mitigates this by allowing clients to train locally and send only model updates to a central server for aggregation~\cite{li2020review} (Figure~\ref{fig:learning_para_stru}(D)).

Traditional FL, however, still faces issues such as fault tolerance, privacy leakage, and high communication cost. To address these, MRAR~\cite{xu2023ring}, a representative form of \textbf{\Acl{DFL} (\acs{DFL})} replaces the central server with a ring all-reduce topology (Figure \ref{fig:learning_para_stru}(E)), where devices exchange parameters with neighbors to collaboratively train a global model. Although \DFL improves robustness by removing the central aggregator, it still lacks incentives and trust mechanisms among potentially untrusted participants. These challenges motivate \emph{fully decentralized learning frameworks} (Figure~\ref{fig:learning_para_stru}(F)), which aim to eliminate single points of failure and support verifiable coordination.

\subsection{Blockchain and Blockchain-based \DeAI}

Blockchain is a distributed ledger technology that records transactions across a decentralized network without relying on a central entity~\cite{Nakamoto2008Bitcoin}. Transactions are batched into blocks, which are cryptographically linked to form an immutable chain. Consensus mechanisms~\cite{bano2019sok} enable participants to agree on the ledger's state. Smart contracts are programs stored on the blockchain that will be executed when predefined conditions are met, and they are widely applied across many domains~\cite{werner2022sok, jiang2023decentralized, saberi2019blockchain}.

Blockchain's properties can be applied to address the limitations of \CeAI~\cite{Chandran2025BlockchainDL}. For instance, as shown in Figure~\ref{fig:learning_para_stru}(F), blockchain can support decentralized model training that reduces centralized control and enhances trust in \AI systems~\cite{Xie2024BlockchainIoT,Fernandez2018BlockchainIoT,Chen2018BlockchainTrust}. Moreover, blockchains' native cryptocurrency enables reward mechanisms to incentivize data and model contributors, facilitating sustainable participation in AI ecosystems~\cite{Swan2015Blockchain}. Although numerous ongoing academic and industry works~\cite{Zhang2021BlockchainAI,BittensorWhitepaper,flock-white-paper,cheng2024oml} aim to leverage blockchain to build robust \DeAI systems, blockchain-based \DeAI still faces challenges such as scalability, performance, privacy, and security risks~\cite{Xu2019Scalable,Shrestha2020Privacy,Miller2018Challenges}. 

\begin{figure*}[t]
\centering
\includegraphics[width=\linewidth]{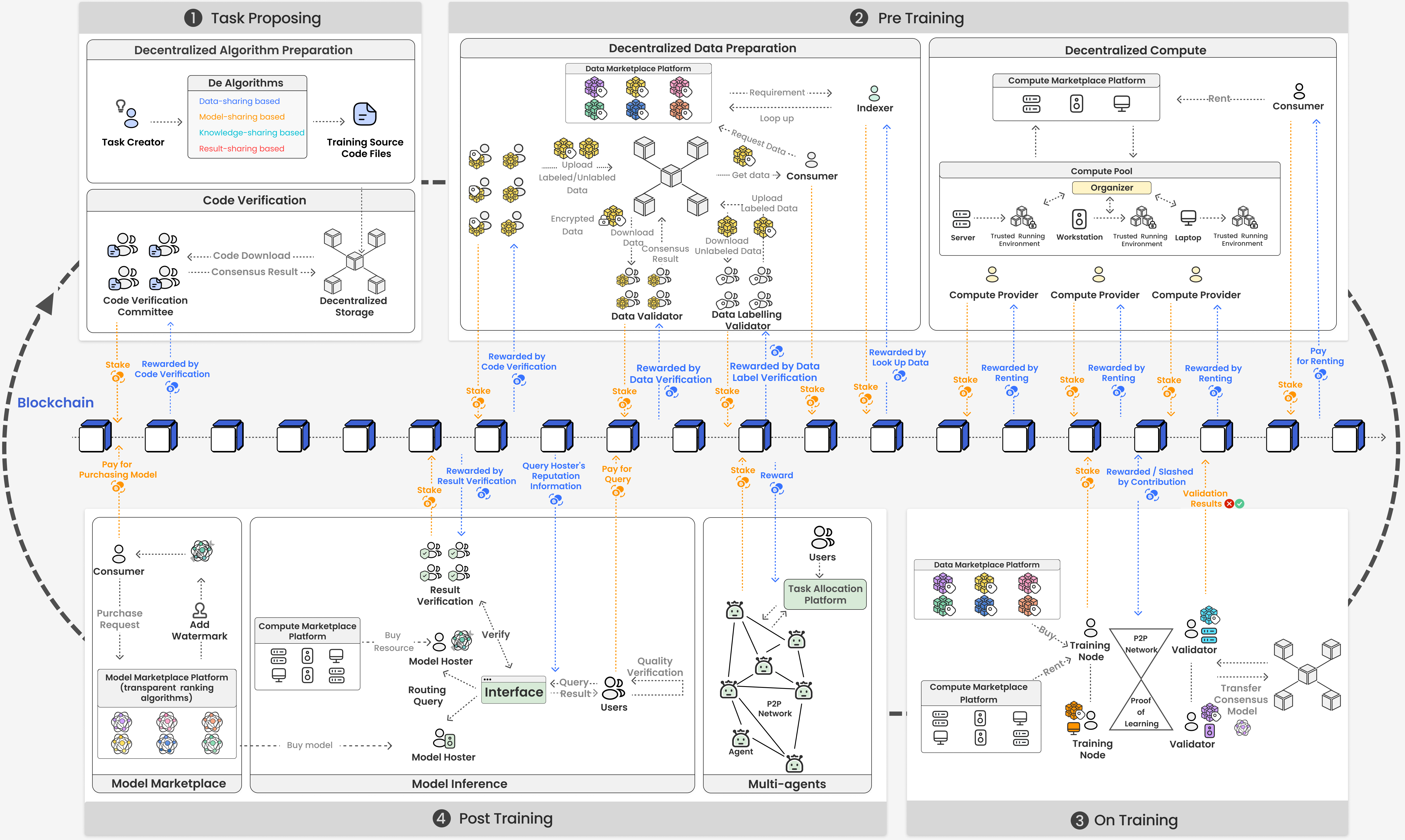}
\caption{A \DeAI model lifecycle  consists of four phases: \protect\blackCircledNum{1} task proposing, \protect\blackCircledNum{2} pre-training, \protect\blackCircledNum{3} on-training, and \protect\blackCircledNum{4} post-training.}
\label{fig:overall_stru}
\end{figure*}

\subsection{Comparison to Existing Work}

Research at the intersection of decentralization, learning, and blockchain spans multiple communities, and several surveys have examined parts of this space. Prior works study classical decentralized learning~\cite{lian2017decentralized,hegedus2019gossip,Allouah2023Mixing,Karimireddy2021Bucketing,Allouah2023Heterogeneous,Gorbunov2023Variance}, decentralized federated learning~\cite{Ehsan2024DFLsurvey,Liangqi2024DFLsurvey,Martinez2023DFLsurvey,Gabrielli2023DFLsurvey,Leon2022IncentivizedFL}, and blockchain-enhanced \ML~\cite{Kayikci2024BlockchainML,Ural2023BlockchainEnhancedML,Khaled2019BlockchainAI,Adedoyin2021BlockchainAI}. However, despite their algorithmic strengths, these non-blockchain decentralized methods generally lack built-in primitives for \emph{trust}, \emph{incentive alignment}, and \emph{verifiable coordination} among mutually distrustful participants. This gap motivates growing interest in blockchain-backed \DeAI, where blockchain and smart contracts can furnish tamper-resistant execution, transparent auditability, and automated incentive mechanisms. Consequently, while \DeAI is much richer than blockchain-based approaches alone, this survey focuses specifically on blockchain-enabled \DeAI due to its unique capabilities in addressing trust and incentive challenges in open, adversarial environments.

Complementary lines survey blockchain-based \ML ~\cite{Li2022BCFL,Wang2024BlockchainedFL,Ning2024BFLSurvey,Wu2023BFLSurvey,Cai2024BCFL}. However, these works generally treat blockchain as an auxiliary component for FL or ML, do not analyze the economic and incentive-driven aspects of permissionless \DeAI, and lack empirical evaluation of deployed platforms such as Bittensor~\cite{BittensorWhitepaper} and FLock~\cite{flock-white-paper}. Our work differs by providing a blockchain-centered threat model, a cross-layer taxonomy, and an empirical analysis of real-world \DeAI systems. A detailed comparison with prior surveys is provided in Appendix~\ref{app:related-work}.

\section{\DeAI Framework}
\label{sec:deai-training}

In this section, in response to \textbf{RQ1}, we first propose a formalization of a \DeAI system, to be followed by a general taxonomy of \DeAI across various ML lifecycles. We conclude with a series of decentralization metrics for \DeAI.  

\subsection{\DeAI Formalization}
\label{sec:deai-formalization}
We formalize a \DeAI system as a tuple
$\mathcal{S} = \bigl(M,\;G,\;\{D_i\}_{i=1}^M,\;\Theta,\;P,\;\Pi,\;\Gamma,\;V_{\mathrm{val}},\;D_{\mathrm{del}},\;\delta\bigr)$, where:
\begin{itemize}[leftmargin=*]
  \item $M$ is the set of \emph{miners} who contribute directly to data, compute resources, model training or inference.
  \item $G = (M, E, W)$ is the \Acl{PtoP} (\acs{PtoP}) communication graph over miners, with adjacency matrix $W\in\mathbb{R}^{|M|\times |M|}$.
  \item $D_i$ is the private data distribution held locally by miner~$i\in M$.
  \item $\Theta$ is the shared parameter space; each miner $i$ maintains a local copy $\theta_i\in\Theta$.
  \item $P:\mathcal{T}\to\mathcal{O}$ is the \emph{task-proposing mechanism} mapping submitted tasks $t\in\mathcal{T}$ to optimization objectives $\mathcal{O}$.
  \item $\Pi$ is the \emph{incentive mechanism}, allocating economic returns across data, compute, training, validation, or delegation contributions.
  \item $\Gamma$ is the \emph{governance function}, which decides protocol updates, task allocations, validator selection, and dispute resolution across all stages.
  \item $V_{\mathrm{val}}$ is the set of \emph{validators} who verify contributions (e.g., model updates, inference proofs) from $M$ and commit checkpoints on-chain.
  \item $D_{\mathrm{del}}$ is the set of \emph{delegators}, who lock stake and delegate it to $M$ or $V_{\mathrm{val}}$.
  \item $\delta: D_{\mathrm{del}}\to{2^{M}\cup 2^{V_{\mathrm{val}}}}$ is the delegation mapping specifying which $M$ or $V_{\mathrm{val}}$ each delegator backs.
\end{itemize}

This tuple provides a unified abstraction for the \DeAI lifecycle. Different stages activate different components: task proposing operates primarily through $P$ and $\Gamma$, pre-training leverages $\{D_i\}$ and compute allocation over $G$, on-training instantiates iterative gradient updates and validation, while post-training invokes $\Pi$, $V_{\mathrm{val}}$, and $\delta$ to ensure attribution, rewards, and on-chain auditability. 

As a concrete instantiation, consider the training stage. Each miner $i \in M$ performs:
\begin{itemize}[leftmargin=*]
  \item \emph{Local compute}: compute 
  $L_i(\theta_i) = \mathbb{E}_{x\sim D_i}[\ell(\theta_i; x)]$, then update
  \[
    \theta_i^{(k+\tfrac12)} 
    = \theta_i^{(k)} - \eta\,\nabla L_i\!\bigl(\theta_i^{(k)}\bigr).
  \]
 
  \item \emph{Neighbor exchange}: share $\theta_i^{(k+\tfrac12)}$ with peers according to the graph $G$.
  
  \item \emph{Local mix}: aggregate neighbor updates via
  \[
    \theta_i^{(k+1)} = \sum_{j \in M} W_{ij}\,\theta_j^{(k+\tfrac12)}.
  \]
\end{itemize}

Under mild conditions on $W$ (e.g., doubly-stochastic, connected) and $\ell(\cdot;x)$ (e.g., smoothness, convexity), the local parameters $\theta_i^{(k)}$ converge to a common optimum
\[
  \theta^* = \arg\min_{\theta\in\Theta}\sum_{i\in M} L_i(\theta),
\]
without centralizing data or parameters. Note that $G$ is of particular importance here: in a centralized setting, every worker communicates with a single parameter server, whereas in \DeAI there is no central coordinator. Instead, to finalize progress, validators $V_{\mathrm{val}}$ checkpoint the agreed parameters on-chain, while Delegators strengthen this process by staking to selected miners $M$ or validators $V_{\mathrm{val}}$ via $\delta$, thereby securing the protocol and receiving proportional rewards through $\Pi$.

\subsection{\DeAI Lifecycle}

We now describe how the abstract system $\mathcal{S}$ unfolds in practice. As shown in Figure~\ref{fig:overall_stru}, the \DeAI lifecycle comprises four phases: task proposing, pre-training, on-training, and post-training, with a feedback loop that may return to task proposal for refinement.

\noindent \blackCircledNum{1}~\textbf{Task proposing} is captured by $P$ and $\Gamma$: nodes submit candidate tasks, which are validated and broadcast over $G$. See Appendix \ref{app:task-proposing} for detailed explanations.

\noindent \blackCircledNum{2}~\textbf{Pre-training} engages $\{D_i\}$ for data preparation and compute allocation over $G$.  

\noindent \blackCircledNum{3}~\textbf{On-training} instantiates the gradient–exchange–mix dynamics of Sec.~\ref{sec:deai-formalization}.

\noindent \blackCircledNum{4}~\textbf{Post-training} invokes $\Pi$ and $\delta$ to  distribute rewards, and enable downstream applications such as inference, agents, and model marketplaces.  

This mapping ensures that each lifecycle phase is grounded in the formal system $\mathcal{S}$ while highlighting their distinct operational roles.

\subsection{Decentralization Metrics for \DeAI}
We evaluate a \DeAI system with the following metrics:

\begin{itemize}[leftmargin=*]
  \item \emph{Data:}  
    Each $D_i$ remains local, with no central data pool.

  \item \emph{Architecture \& Compute:}  
    The communication graph $G$ (and its weight matrix $W$) defines the \PtoP compute topology.

    \item \emph{Model Ownership:}  
    A mapping $O:\Theta \to V$ tracks which node owns each parameter instance $\theta \in \Theta$, with ownership enforced via smart contracts.

  \item \emph{Incentive Mechanism:}  
    The incentive function $\Pi$ rewards contributions without centralized control.

  \item \emph{Governance:}  
    Protocol upgrades, task allocation, and dispute resolution are governed by a decentralized protocol $\Gamma$.
\end{itemize}

In practice, we consider a system to qualify as \DeAI if it satisfies some or all of the above metrics, with stronger decentralization achieved as more dimensions are met.

\subsection{Blockchain Functionalities in \DeAI}

Blockchain offers diverse functionalities that enhance the security, transparency, and trustworthiness of \DeAI processes. These can be summarized as follows:

\begin{itemize}

    \item \emph{Incentive Mechanisms}:  
    Native token economies incentivize participation by rewarding various contributions in a decentralized system. This ensures fair compensation and fosters sustainable collaboration in \DeAI ecosystems.

    \item \emph{Governance and Access Control}:  
    Permissionless consensus and smart contracts enable decentralized governance and trustless collaboration. Access to data, compute resources and models is managed without central authorities, with agreements automatically enforced.

    \item \emph{Auditability and Transparency}:  
    The immutable blockchain ledger provides full traceability of contribution as well as usage of data, compute, models, and inferences. This ensures accountability, supports regulatory compliance, and strengthens trust in \DeAI outcomes.


    \item \emph{Security}:  
    Blockchain offers tamper-resistant storage, strong provenance guarantees, and resilience against single points of failure, supporting trustworthy coordination in decentralized environments.

\end{itemize}

While blockchain provides useful security properties, it is not a universal solution for securing \DeAI or \AI more broadly. Decentralization does not automatically imply stronger security: heterogeneous participants may have uneven protection levels, and thus expanding the attack surface. Immutability aids auditing, but cannot by itself prevent poisoning or other Byzantine behaviors without complementary detection or robust aggregation. Transparency also creates privacy–auditability trade-offs that require additional techniques. Thus, blockchain’s security benefits depend on system design, threat models, and integration with other protections. Section~\ref{sec:security_risks} expands on these risks and mitigations.

In the sections that follow, we examine blockchain-based \DeAI across pre-training, training, and post-training. For each stage, we address \textbf{RQ2} by analyzing blockchain’s role in decentralization and security, and \textbf{RQ3} by distilling insights and research gaps arising from current industry implementations.

\begin{table*}[!htbp]
\caption{Overview of \DeAI Projects}
\label{tab:deai_blockchain}
\begin{threeparttable}
    \centering
\renewcommand\arraystretch{1.03}
\resizebox{\linewidth}{!}{
\begin{tabular}{l|ccccccc|ccccccc|c|c|l}
\multirow{2}{*}{\textbf{Projects}} & \multicolumn{7}{c|}{\textbf{\DeAI Framework}} & \multicolumn{7}{c|}{\textbf{Blockchain Functionalities}} & \multirow{2}{*}{\rotatebox{90}{Decentralization$\ssymbol{2}$}} & \multirow{2}{*}{\rotatebox{90}{Staking}} & \multirow{2}{*}{\rotatebox{0}{\makecell{Security Guarantee\\ Solutions}}} \\
\cline{2-15}
  & \rotatebox{90}{\makecell{Task\\ Creation}} & \rotatebox{90}{\makecell{Data\\ Preparation}} & \rotatebox{90}{\makecell{Compute}} & \rotatebox{90}{\makecell{Training}} & \rotatebox{90}{\makecell{Model\\ Inference}} & \rotatebox{90}{\makecell{Model\\ Marketplace}} & \rotatebox{90}{Agents} & \rotatebox{90}{\makecell{Incentive\\ Mechanism}} & \rotatebox{90}{\makecell{Enhanced\\ Security}} & \rotatebox{90}{\makecell{Permission \\Control}} & \rotatebox{90}{\makecell{Data\\ Storage}} & \rotatebox{90}{\makecell{Public\\ Reference}} & \rotatebox{90}{Auditability} & \rotatebox{90}{\makecell{\AI Assets\\ Tokenization}} & & & \\ 
\midrule

\cellcolor{brown!15}
Vana~\cite{vana-docs} &\whitecircle &\blackcircle &\whitecircle &\whitecircle &\whitecircle &\whitecircle &\whitecircle
&\blackcircle &\blackcircle &\blackcircle &\blackcircle &\blackcircle &\blackcircle &\blackcircle
&\halfblackcircle &\blackcircle &\ZKP \\

\cellcolor{blue!10}
Fraction AI~\cite{fractionai-docs} &\whitecircle &\blackcircle &\whitecircle &\whitecircle &\whitecircle &\whitecircle &\whitecircle
&\blackcircle &\blackcircle &\blackcircle &\blackcircle &\blackcircle &\blackcircle &\blackcircle
&\halfblackcircle &\blackcircle &Reputation\\

\cellcolor{brown!15}
Ocean~\cite{ocean-white-paper} &\whitecircle &\blackcircle &\whitecircle &\whitecircle &\whitecircle &\whitecircle &\whitecircle
&\blackcircle &\blackcircle &\blackcircle &\blackcircle &\blackcircle &\blackcircle &\blackcircle
&\halfblackcircle &\blackcircle &{On-chain Consensus} \\

\cellcolor{blue!10}
Numbers~\cite{numbers-white-paper} &\whitecircle &\blackcircle &\whitecircle &\whitecircle &\whitecircle &\whitecircle &\whitecircle
&\blackcircle &\blackcircle &\blackcircle &\blackcircle &\blackcircle &\blackcircle &\blackcircle
&\halfblackcircle &\blackcircle &\PoS \\

\cellcolor{brown!15}
The Graph~\cite{the-graph-white-paper} &\whitecircle &\blackcircle &\whitecircle &\whitecircle &\whitecircle &\whitecircle &\whitecircle
&\blackcircle &\blackcircle &\blackcircle &\blackcircle &\blackcircle &\blackcircle &\blackcircle
&\halfblackcircle &\blackcircle &{On-chain Consensus} \\

\cellcolor{blue!10}
Synternet~\cite{synternet-docs} &\whitecircle &\blackcircle &\whitecircle &\whitecircle &\whitecircle &\whitecircle &\whitecircle
&\blackcircle &\blackcircle &\blackcircle &\blackcircle &\blackcircle &\blackcircle &\blackcircle
&\halfblackcircle &\blackcircle &Proof-of-Delivery/Consumption \\

\cellcolor{brown!15}
OriginTrail~\cite{origin-trail-white-paper} &\whitecircle &\blackcircle &\whitecircle &\whitecircle &\whitecircle &\whitecircle &\whitecircle
&\blackcircle &\blackcircle &\blackcircle &\blackcircle &\blackcircle &\blackcircle &\blackcircle
&\halfblackcircle &\blackcircle &Proof-of-Knowledge\\

\cellcolor{blue!10}
ZeroGravity~\cite{zero-gravity-white-paper} &\whitecircle &\blackcircle &\whitecircle &\whitecircle &\whitecircle &\whitecircle &\whitecircle
&\blackcircle &\blackcircle &\blackcircle &\blackcircle &\blackcircle &\blackcircle &\blackcircle
&\halfblackcircle &\blackcircle & Proof-of-Random Access \\

\cellcolor{brown!15}
Grass~\cite{grass-docs} &\whitecircle &\blackcircle &\whitecircle &\whitecircle &\whitecircle &\whitecircle &\whitecircle
&\blackcircle &\blackcircle &\blackcircle &\blackcircle &\blackcircle &\blackcircle &\blackcircle
&\halfblackcircle &\blackcircle &ZKP + Reputation \\

\cellcolor{blue!10}
OORT Storage~\cite{oort-docs} &\whitecircle &\blackcircle &\whitecircle &\whitecircle &\whitecircle &\whitecircle &\whitecircle
&\blackcircle &\blackcircle &\blackcircle &\blackcircle &\blackcircle &\blackcircle &\blackcircle
&\halfblackcircle &\blackcircle &Proof-of-Honesty\\

\cellcolor{brown!15}
KIP~\cite{kip-docs} &\whitecircle &\blackcircle &\whitecircle &\whitecircle &\whitecircle &\whitecircle &\whitecircle
&\blackcircle &\blackcircle &\blackcircle &\blackcircle &\blackcircle &\blackcircle &\blackcircle
&\halfblackcircle &\halfblackcircle &{On-chain Consensus} \\

\cellcolor{blue!10}
Filecoin~\cite{filecoin-docs} &\whitecircle &\blackcircle &\whitecircle &\whitecircle &\whitecircle &\whitecircle &\whitecircle
&\blackcircle &\blackcircle &\blackcircle &\blackcircle &\blackcircle &\blackcircle &\blackcircle
&\blackcircle &\whitecircle &Proof-of-Replication/Spacetime \\

\cellcolor{blue!10}
OpenLedger~\cite{openledger-white-paper} &\whitecircle &\blackcircle &\whitecircle &\blackcircle &\blackcircle &\blackcircle &\whitecircle
&\blackcircle &\blackcircle &\blackcircle &\blackcircle &\blackcircle &\blackcircle &\blackcircle
&\blackcircle &\blackcircle &Proof-of-Attribution \\

\hline


\cellcolor{brown!15}
IO.NET~\cite{io-net-docs} &\whitecircle &\whitecircle &\blackcircle &\whitecircle &\whitecircle &\whitecircle &\whitecircle
&\blackcircle &\blackcircle &\blackcircle &\whitecircle &\blackcircle &\blackcircle &\blackcircle
&\halfblackcircle &\blackcircle & Reward + Slash \\

\cellcolor{blue!10}
NetMind~\cite{netmind-white-paper} &\whitecircle &\whitecircle &\blackcircle &\whitecircle &\whitecircle &\whitecircle &\whitecircle
&\blackcircle &\blackcircle &\blackcircle &\whitecircle &\blackcircle &\blackcircle &\blackcircle
&\halfblackcircle &\blackcircle &Proof-of-Authority + MPC\\

\cellcolor{brown!15}
Render Network~\cite{render-network-docs} &\whitecircle &\whitecircle &\blackcircle &\whitecircle &\whitecircle &\whitecircle &\whitecircle
&\blackcircle &\blackcircle &\blackcircle &\whitecircle &\blackcircle &\blackcircle &\blackcircle
&\halfblackcircle &\blackcircle &Reputation + Proof-of-Render \\

\cellcolor{blue!10}
Akash~\cite{akash-white-paper} &\whitecircle &\whitecircle &\blackcircle &\whitecircle &\whitecircle &\whitecircle &\whitecircle
&\blackcircle &\blackcircle &\blackcircle &\whitecircle &\blackcircle &\blackcircle &\blackcircle
&\halfblackcircle &\blackcircle &Tendermint Consensus \\

\cellcolor{brown!15}
Nosana~\cite{nosana-docs} &\whitecircle &\whitecircle &\blackcircle &\whitecircle &\whitecircle &\whitecircle &\whitecircle
&\blackcircle &\blackcircle &\blackcircle &\whitecircle &\blackcircle &\blackcircle &\blackcircle
&\halfblackcircle &\blackcircle &{On-chain Consensus} \\


\cellcolor{blue!10}
Inferix~\cite{inferix-white-paper} &\whitecircle &\whitecircle &\blackcircle &\whitecircle &\whitecircle &\whitecircle &\whitecircle
&\blackcircle &\blackcircle &\blackcircle &\whitecircle &\blackcircle &\blackcircle &\blackcircle
&\halfblackcircle &\blackcircle &Proof-of-Rendering \\

\cellcolor{brown!15}
OctaSpace~\cite{octaSpace-white-paper} &\whitecircle &\whitecircle &\blackcircle &\whitecircle &\whitecircle &\whitecircle &\whitecircle
&\blackcircle &\blackcircle &\blackcircle &\whitecircle &\blackcircle &\blackcircle &\blackcircle
&\halfblackcircle &\blackcircle &{On-chain Consensus} \\

\cellcolor{blue!10}
DeepBrain Chain~\cite{deep-brain-chain-white-paper} &\whitecircle &\whitecircle &\blackcircle &\whitecircle &\whitecircle &\whitecircle &\whitecircle
&\blackcircle &\blackcircle &\blackcircle &\whitecircle &\blackcircle &\blackcircle &\blackcircle
&\halfblackcircle &\blackcircle &Delegated \PoS \\

\cellcolor{brown!15}
OpSec~\cite{opsec-docs} &\whitecircle &\whitecircle &\blackcircle &\whitecircle &\whitecircle &\whitecircle &\whitecircle
&\blackcircle &\blackcircle &\blackcircle &\whitecircle &\blackcircle &\blackcircle &\blackcircle
&\halfblackcircle &\blackcircle &Delegated \PoS \\

\cellcolor{blue!10}
Gensyn~\cite{gensyn-lite-paper} &\whitecircle &\whitecircle &\blackcircle &\whitecircle &\whitecircle &\whitecircle &\whitecircle
&\blackcircle &\blackcircle &\blackcircle &\whitecircle &\blackcircle &\blackcircle &\blackcircle
&\halfblackcircle &\blackcircle &\Acl{PoL} (\acs{PoL}) \\

\cellcolor{brown!15}

Lilypad~\cite{Eisele2020Lilypad} &\whitecircle &\whitecircle &\blackcircle &\whitecircle &\whitecircle &\whitecircle &\whitecircle
&\blackcircle &\blackcircle &\blackcircle &\whitecircle &\blackcircle &\blackcircle &\blackcircle
&\halfblackcircle &\blackcircle &Mediators + On-chain consensus\\

\hline

\cellcolor{blue!10}
Bittensor~\cite{rao2020bittensor} &\halfblackcircle &\whitecircle &\whitecircle &\blackcircle &\whitecircle &\whitecircle &\whitecircle
&\blackcircle &\blackcircle &\blackcircle &\whitecircle &\blackcircle &\blackcircle &\blackcircle
&\halfblackcircle &\blackcircle &Yuma Consensus \\ 

\cellcolor{brown!15}
FLock.io~\cite{flock-white-paper} &\halfblackcircle &\whitecircle &\whitecircle &\blackcircle &\blackcircle &\whitecircle &\whitecircle
&\blackcircle &\blackcircle &\blackcircle &\whitecircle &\blackcircle &\blackcircle &\blackcircle
&\halfblackcircle &\blackcircle &FLock Consensus \\

\cellcolor{blue!10}
Numerai~\cite{numerai-docs} &\whitecircle &\whitecircle &\whitecircle &\blackcircle &\blackcircle &\whitecircle &\whitecircle
&\blackcircle &\blackcircle &\blackcircle &\whitecircle &\blackcircle &\blackcircle &\blackcircle
&\halfblackcircle &\halfblackcircle &{On-chain Consensus} \\

\cellcolor{brown!15}
Commune AI~\cite{commune-ai-docs} &\halfblackcircle &\whitecircle &\whitecircle &\blackcircle &\whitecircle &\whitecircle &\whitecircle
&\blackcircle &\blackcircle &\blackcircle &\whitecircle &\blackcircle &\blackcircle &\blackcircle
&\halfblackcircle &\blackcircle &Yuma Consensus \\
\hline

\cellcolor{blue!10}
Modulus~\cite{modulus-white-paper} &\whitecircle &\whitecircle &\whitecircle &\whitecircle &\blackcircle &\whitecircle &\whitecircle
&\blackcircle &\blackcircle &\blackcircle &\whitecircle &\blackcircle &\blackcircle &\blackcircle
&\halfblackcircle &\whitecircle &zkML \\

\cellcolor{brown!15}
Hyperspace~\cite{hyperspace-white-paper} &\whitecircle &\whitecircle &\whitecircle &\whitecircle &\blackcircle &\whitecircle &\whitecircle
&\blackcircle &\blackcircle &\blackcircle &\whitecircle &\blackcircle &\blackcircle &\blackcircle
&\halfblackcircle &\whitecircle &Fraud Proof \\

\cellcolor{blue!10}
Sertn~\cite{sertn-white-paper} &\whitecircle &\whitecircle &\whitecircle &\whitecircle &\blackcircle &\whitecircle &\whitecircle
&\blackcircle &\blackcircle &\blackcircle &\whitecircle &\blackcircle &\blackcircle &\blackcircle
&\halfblackcircle &\halfblackcircle &ZKP+FHE$\ssymbol{3}$+MPC \\

\cellcolor{brown!15}
ORA~\cite{ora-white-paper} &\whitecircle &\whitecircle &\whitecircle &\whitecircle &\blackcircle &\whitecircle &\whitecircle
&\blackcircle &\blackcircle &\blackcircle &\whitecircle &\blackcircle &\blackcircle &\blackcircle
&\halfblackcircle &\blackcircle &opML \\

\cellcolor{blue!10}

Ritual~\cite{rituals-docs} &\whitecircle &\whitecircle &\whitecircle &\whitecircle &\blackcircle &\whitecircle &\whitecircle
&\blackcircle &\blackcircle &\blackcircle &\whitecircle &\blackcircle &\blackcircle &\blackcircle
&\halfblackcircle &\blackcircle &On-chain Consensus \\

\cellcolor{brown!15}

Allora~\cite{allora-white-paper} &\whitecircle &\whitecircle &\whitecircle &\whitecircle &\blackcircle &\whitecircle &\whitecircle
&\blackcircle &\blackcircle &\blackcircle &\whitecircle &\blackcircle &\blackcircle &\blackcircle
&\halfblackcircle &\blackcircle &CometBFT \\
\hline

\cellcolor{blue!10}
Fetch.AI~\cite{fetch-ai-white-paper} &\whitecircle &\whitecircle&\whitecircle &\whitecircle &\whitecircle &\whitecircle &\blackcircle
&\blackcircle &\blackcircle &\blackcircle &\whitecircle &\blackcircle &\blackcircle &\blackcircle
&\halfblackcircle &\blackcircle &\PoS \\

\cellcolor{brown!15} 
Arbius~\cite{arbius-white-paper} &\whitecircle &\whitecircle &\whitecircle &\whitecircle &\whitecircle &\blackcircle &\blackcircle
&\blackcircle &\blackcircle &\blackcircle &\whitecircle &\blackcircle &\blackcircle &\blackcircle
&\halfblackcircle &\blackcircle &\Acl{uPoW} (\acs{uPoW}) \\

\cellcolor{blue!10}
Theoriq~\cite{theoriq-white-paper} &\whitecircle &\whitecircle &\whitecircle &\whitecircle &\whitecircle &\whitecircle &\blackcircle
&\blackcircle &\blackcircle &\blackcircle &\whitecircle &\blackcircle &\blackcircle &\blackcircle
&\halfblackcircle &\blackcircle &Proof-of-Contribution/Collaboration \\

\cellcolor{brown!15}
Delysium~\cite{delysium-white-paper} &\whitecircle &\whitecircle &\whitecircle &\whitecircle &\whitecircle &\whitecircle &\blackcircle
&\blackcircle &\blackcircle &\blackcircle &\whitecircle &\blackcircle &\blackcircle &\blackcircle
&\halfblackcircle &\blackcircle &{On-chain Consensus} \\

\cellcolor{blue!10}
OpenServ~\cite{open-serv-white-paper} &\whitecircle &\whitecircle &\whitecircle &\whitecircle &\whitecircle &\whitecircle &\blackcircle
&\blackcircle &\blackcircle &\blackcircle &\whitecircle &\blackcircle &\blackcircle &\blackcircle
&\halfblackcircle &\whitecircle &{On-chain Consensus} \\

\cellcolor{brown!15}
Autonolas~\cite{autonolas-white-paper} &\whitecircle &\whitecircle &\whitecircle &\whitecircle &\whitecircle &\whitecircle &\blackcircle
&\blackcircle &\blackcircle &\blackcircle &\whitecircle &\blackcircle &\blackcircle &\blackcircle
&\halfblackcircle &\blackcircle &Tendermint Consensus\\


\cellcolor{blue!10}

ELNA~\cite{elna-white-paper} &\whitecircle &\whitecircle &\whitecircle &\whitecircle &\whitecircle &\whitecircle &\blackcircle
&\blackcircle &\blackcircle &\blackcircle &\whitecircle &\blackcircle &\blackcircle &\blackcircle
&\halfblackcircle &\blackcircle &{On-chain Consensus} \\

\cellcolor{brown!15}
OpenAgents~\cite{open-agents-docs} &\whitecircle &\whitecircle &\whitecircle &\whitecircle &\whitecircle &\blackcircle &\blackcircle
&\blackcircle &\blackcircle &\blackcircle &\whitecircle &\blackcircle &\blackcircle &\blackcircle
&\halfblackcircle &\whitecircle &{On-chain Consensus} \\

\hline

\cellcolor{blue!10}
SingularityNET~\cite{singularity-net-docs} &\whitecircle &\whitecircle &\whitecircle &\whitecircle &\whitecircle &\blackcircle &\whitecircle
&\blackcircle &\blackcircle &\blackcircle &\whitecircle &\blackcircle &\blackcircle &\blackcircle
&\halfblackcircle &\whitecircle &Multi-Party Escrow\\

\cellcolor{brown!15}
SaharaAI~\cite{sahara-ai-docs} &\whitecircle &\whitecircle &\whitecircle &\whitecircle &\whitecircle &\blackcircle &\whitecircle
&\blackcircle &\blackcircle &\blackcircle &\whitecircle &\blackcircle &\blackcircle &\blackcircle
&\halfblackcircle &\blackcircle &Proof-of-Stake \\

\cellcolor{blue!10}
Shinkai~\cite{shinkai-white-paper} &\whitecircle &\whitecircle &\whitecircle &\whitecircle &\whitecircle &\blackcircle &\whitecircle
&\blackcircle &\blackcircle &\blackcircle &\whitecircle &\blackcircle &\blackcircle &\blackcircle
&\halfblackcircle &\blackcircle &\ZKP+MPC \\

\cellcolor{brown!15}
Balance DAO~\cite{balance-dao-docs} &\whitecircle &\whitecircle &\whitecircle &\whitecircle &\whitecircle &\blackcircle &\blackcircle
&\blackcircle &\blackcircle &\blackcircle &\whitecircle &\blackcircle &\blackcircle &\blackcircle
&\halfblackcircle &\blackcircle &\PoS \\


\cellcolor{blue!10}
Immutable Labs~\cite{immutable-labs-GPOW} &\whitecircle &\whitecircle &\whitecircle &\whitecircle &\whitecircle &\blackcircle &\whitecircle
&\blackcircle &\blackcircle &\blackcircle &\whitecircle &\blackcircle &\blackcircle &\blackcircle
&\halfblackcircle &\halfblackcircle &Green \PoW\\





\hline
\cellcolor{red!20}
Prime Intellect$\ssymbol{4}$~\cite{prime-intellect-docs} &\whitecircle &\whitecircle &\blackcircle &\blackcircle &\blackcircle &\whitecircle &\whitecircle
&\whitecircle &\whitecircle &\whitecircle &\whitecircle &\whitecircle &\whitecircle &\whitecircle
&\whitecircle &\whitecircle &Centralized Server \\

\bottomrule
\end{tabular}
}

\begin{tablenotes}
  \scriptsize
  \item[]\quad \blackcircle\ Supported \quad
         \halfblackcircle\ Partially supported \quad
         \whitecircle\ Not supported
  \item[] $\ssymbol{2}$\textbf{Decentralization}: “partially” decentralization means a project has centralized or off-chain components.
  $\ssymbol{3}$\textbf{FHE:} Fully Homomorphic Encryption.
  \item[] $\ssymbol{4}$\textbf{Prime Intellect}: We also present the project which aims to build \DeAI but does not explicitly mention blockchain in its design.
\end{tablenotes}



\end{threeparttable}

\end{table*}

\section{Pre-Training}

\subsection{Data Preparation}

\textbf{Data preparation} involves processes such as collection, cleaning, normalization, transformation, and feature selection. These steps are crucial for effective \AI model training, directly shaping model accuracy, generalization~\cite{garcia2016data} and interpretability~\cite{guyon2003introduction}. As discussed in \S\ref{sec:deai-training}, in the \DeAI tuple $\mathcal{S}$, data preparation corresponds to the distributed datasets $\{D_i\}_{i \in M}$ contributed by heterogeneous miners.

\subsubsection{Challenges of Centralized Data Preparation}
When $D_i$ are centralized into a single pool, scaling becomes constrained. \LLMs demand massive and high-quality datasets to train: GPT-3 required 1.2TB~\cite{Brown2020GPT3}, while the estimated supply of suitable public text is only 6TB~\cite{villalobosposition}. This finite pool suggests limits to further scaling~\cite{longpre2024consent}. Centralized pipelines also risk bias from underrepresented domains and languages, and privacy regulations may restrict access to sensitive datasets (e.g., healthcare). Thus, central data control creates bottlenecks in both scale and representativeness.

\subsubsection{Decentralized Solutions for Data Preparation}
In \DeAI, each miner contributes $D_i$ directly without central pooling, with incentives $\Pi$ and governance $\Gamma$ aligning honest and meaningful participation. Training frameworks such as FL~\cite{dong2023defending} allow secure, privacy-preserving contributions while maintaining ownership. However, the lack of a central curator introduces three risks: \ding{192} malicious or low-quality submissions, \ding{193} free-riding for rewards, and \ding{194} potential leakage of private data. Emerging protocols~\cite{vana-docs, fractionai-docs, ocean-white-paper, numbers-white-paper, the-graph-white-paper, synternet-docs, origin-trail-white-paper, zero-gravity-white-paper, grass-docs, oort-docs, kip-docs, filecoin-docs} mitigate these issues through:
\paragraph{Incentive Mechanisms ($\Pi$)}
\noindent\emph{Dataset Tokenization:} Ocean~\cite{ocean-white-paper} and Vana~\cite{vana-docs} tokenize datasets into datatokens and data providers can earn fees; \emph{Proof-of-Data Contribution}: Vana~\cite{vana-docs} leverages validators $V_{\mathrm{val}}$ to score data quality via influence functions, linking contributor rewards directly to measured utility; \emph{Stake and Reputation}:  Fraction AI~\cite{fractionai-docs} couples staking with contributor reputation, meaning high-reputation miners receive priority and greater rewards, while poor-quality submissions reduce reputation and may be slashed.

\paragraph{Privacy Protection}
To protect $D_i$, Vana integrates \ZKPs~\cite{Groth2016Groth16} for verifiable data integrity without disclosure of the actual data itself~\cite{vana-zkp-docs}. Ocean Protocol~\cite{ocean-white-paper} builds access control through datatokens and smart contracts, ensuring that only authorized nodes access encrypted datasets.

\paragraph{Verification and Provenance ($\Gamma$)}
Blockchain’s immutable ledger allows $\Gamma$ to enforce transparent provenance of $D_i$. Numbers Protocol~\cite{numbers-white-paper} records metadata and ownership histories on-chain, ensuring authenticity and accountability in collaborative pipelines.

\insightbox{Blockchain-enabled data preparation requires:
\ding{192} \emph{on-chain provenance and immutability} to verify the authenticity and integrity of contributed data;
\ding{193} \emph{smart-contract–driven, tokenized incentive mechanisms} to reward high-quality data contributions; and
\ding{194} \emph{privacy-preserving blockchain integrations} to protect sensitive data while enabling verifiable contribution.}


\subsubsection{Discussion}

Despite the existing attempts~\cite{vana-docs, fractionai-docs, ocean-white-paper, numbers-white-paper} to build decentralized data preparation systems, addressing the trade-offs between rewards, privacy, and authenticity in these solutions highlights several pressing research gaps. First, optimizing incentive structures that offer fair rewards without risking inflation or reward dilution remains a challenge, as current mechanisms~\cite{vana-docs, fractionai-docs} may vary widely in effectiveness and scalability. Additionally, ensuring data authenticity through decentralized consensus mechanisms, such as Proof-of-Contribution in Vana~\cite{vana-docs}, may present scalability issues, particularly as data volumes increase and the need for real-time validation grows. Additionally, once data consumers have paid the fees and gained access to the data, they may forward it to other consumers without sharing any rewards with the original data providers. 
\gapbox{How to develop efficient and scalable privacy solutions, and lightweight yet robust consensus mechanisms that collectively balance data contributor incentives, privacy, and authenticity in decentralized ecosystems?}
\vspace{-0.6\baselineskip}
\subsection{Compute}
\textbf{Compute resources} determine the feasibility and performance of \AI training and inference. In the \DeAI tuple $\mathcal{S}$ described in \S\ref{sec:deai-training}, compute corresponds to the graph $G=(M,E,W)$, which defines how nodes exchange workloads, and to the infrastructure that supports secure, verifiable execution.

\subsubsection{Challenges of Centralized Compute}
Centralized compute remains prohibitively expensive, allowing only a few entities to acquire and scale the necessary infrastructure, which concentrates resources and widens access inequality~\cite{Brown2020GPT3}. Cloud rentals partly mitigate this but still impose high costs on independent researchers. Centralized data centers also suffer from inefficiency, with some GPUs idle while others are oversubscribed. In addition, their environmental impact is substantial: large facilities consume millions of liters of water daily and generate emissions projected to rival those of entire nations~\cite{techstory-ai-environment}. These scalability, cost, and sustainability pressures underscore the fragility of heavily centralized compute.


\subsubsection{Decentralized Solutions for Compute}
\DeAI distributes compute workloads over $G$ by leveraging blockchain to coordinate incentives $\Pi$, governance $\Gamma$, and validation $V_{\mathrm{val}}$. Emerging solutions~\cite{Eisele2020Lilypad, io-net-docs, netmind-white-paper, render-network-docs, akash-white-paper, nosana-docs, octaSpace-white-paper, inferix-white-paper, deep-brain-chain-white-paper, opsec-docs, gensyn-lite-paper, commune-ai-docs} include:

\paragraph{Permissionless Access ($G$)}
Blockchains remove intermediaries, allowing compute providers to join directly as miners in $G$. Lilypad~\cite{Eisele2020Lilypad} executes containerized workloads on idle machines, while Render Network~\cite{render-network-docs} allocates GPU power dynamically across a \PtoP graph.

\paragraph{Incentive Mechanisms ($\Pi$)}
Tokenomics ensure providers are rewarded fairly. IO.NET~\cite{io-net-docs} creates a marketplace for GPU rental, while Akash~\cite{akash-white-paper} ties compute rewards to a \PoS consensus. Delegators $D_{\mathrm{del}}$ can back reliable providers, amplifying incentives for honest participation.

\paragraph{Scalability of Compute Networks ($W$)}
As more miners contribute, decentralized networks expand capacity by aggregating distributed GPUs. Lilypad~\cite{Eisele2020Lilypad} scales horizontally, and Render~\cite{render-network-docs} dynamically allocates workloads via $W$, ensuring balanced task distribution.

\paragraph{Task Verification ($V_{\mathrm{val}}, \Gamma$)}
Validators audit computations to ensure integrity. Render~\cite{render-network-docs} employs an on-chain reputation system, while Gensyn~\cite{gensyn-lite-paper} uses staking plus graph-based protocols to verify task completion. Governance rules $\Gamma$ enforce accountability and resolve disputes.

\paragraph{Security and Integrity}
Protocols such as NetMind~\cite{netmind-white-paper} use multi-party computation (MPC) for secure task execution, while Gensyn~\cite{gensyn-lite-paper} and Akash~\cite{akash-white-paper} enforce honest behavior through staking and Tendermint consensus~\cite{buchman2016tendermint}. These mechanisms protect the system from tampering and ensure verifiable execution.


\insightbox{
Blockchain-supported decentralized computing requires:
\ding{192} \emph{permissionless participation secured by consensus} to allow open access to compute tasks;
\ding{193} \emph{smart-contract–coordinated incentives} to reward honest compute execution;
\ding{194} \emph{scalable off-chain computation frameworks} anchored to the blockchain for verification; and
\ding{195} \emph{verification and attestation mechanisms} to ensure integrity, privacy, and resistance to malicious workers.
}

\subsubsection{Discussion} Staking mechanisms contribute significantly to the security and reliability of decentralized compute. It also serves as a signal of demand in decentralized systems. However, it is important to note that staking is not the only way to achieve these goals. For instance, Lilypad~\cite{Eisele2020Lilypad}, instead of using \PoS, opts for a \uPoW consensus mechanism. Specifically, nodes must be online for a minimum of four hours a day continuously in order to be eligible for rewards. Furthermore, across different protocols for decentralized compute, various tokenomics models are employed. For example, Akash~\cite{akash-white-paper} uses an inflationary model, while IO.NET~\cite{io-net-docs} adopts a disinflationary approach. Specifically, it reduces emitted rewards each month after the first year, whereas Akash’s inflationary model starts at an inflation rate of $100\%$ that halves every two years. In theory, an inflationary model can effectively incentivize compute providers during the early stages of network development, whereas a disinflationary model may help sustain the network's long-term economic health. An open question is to analyze the empirical impacts of these tokenomics models on decentralized compute networks, particularly given the fluctuating costs of compute resources.

\gapbox{What are the empirical impacts of inflationary versus deflationary tokenomics models on decentralized compute networks?}
\vspace{-0.6\baselineskip}
\section{On-Training}

In our formalization of \DeAI (\S\ref{sec:deai-training}), training proceeds through repeated cycles of local compute, neighbor exchange, and local mixing, with the training conducted by $M$ over the communication graph $G$, while validators $V_{\mathrm{val}}$ audit and finalize updates on-chain, and delegators $D_{\mathrm{del}}$ strengthen consensus by staking to miners and/or validators and sharing in the rewards. We now contrast how this process manifests in centralized versus decentralized settings.

\subsection{Challenges of Centralized Training}
In a centralized setup, all nodes exchange parameters only with a central parameter server. Here, the incentive mechanism $\Pi$ and the governance function $\Gamma$ effectively collapse under the control of a single provider. As such, training data $\{D_i\}$ must be uploaded to the central entity, which controls hyperparameter tuning, monitoring, and evaluation~\cite{Deng2009ImageNet}. This centralized instantiation raises several weaknesses: (i) all data is exposed to the central server, threatening privacy and security, especially under regulations such as HIPAA in healthcare~\cite{HIPAA1996}; (ii) only resource-rich entities can train frontier-scale models~\cite{Brown2020GPT3}; (iii) the parameter server becomes a single point of failure, vulnerable to outages or attacks; and (iv) trust and transparency are limited~\cite{Brundage2018Malicious}.

\subsection{Decentralized Solutions for Training}
\DeAI training replaces the single coordinator with \PtoP coordination over $G$, supported by $\Gamma$ for governance and $\Pi$ for incentives. In practice, these design elements manifest along three principles:

\begin{itemize}[leftmargin=*]
\item \textbf{Trustless and Transparent Training:}
With $\Gamma$ encoded on-chain, training progresses without relying on a central authority. For instance, protocols like Bittensor~\cite{BittensorWhitepaper} records rules and contributions transparently, while Numerai~\cite{numerai-docs} aggregates independent participant submissions into a meta-model, ensuring that outcomes are verifiable and auditable.

\item \textbf{Decentralized Model Validation:}
Validators $V_{\mathrm{val}}$ enforce quality control. FLock.io~\cite{flock-white-paper,wang2025aiarena} distributes evaluation datasets from task creators and rewards validators based on score accuracy and staked tokens, aligning incentives with honest behavior without central evaluators.

\item \textbf{Consensus and Incentive Mechanisms:}
The incentive layer $\Pi$ ensures sustained participation. In Bittensor, Yuma consensus~\cite{BittensorWhitepaper} allocates rewards to miners and validators by contribution, while FLock.io ties training and validation rewards to both performance and stake. These mechanisms incentivize meaningful updates and discourage free-riding.
\end{itemize}


\insightbox{
Blockchain-enabled decentralized training requires:
\ding{192} \emph{on-chain task registration} for transparent, verifiable job specification;
\ding{193} \emph{trustless coordination via smart contracts} in place of central orchestrators;
\ding{194} \emph{blockchain-anchored validation} to detect poisoning and ensure correctness;
\ding{195} \emph{token-based incentives} to reward good updates and penalize misuse; and
\ding{196} \emph{on-chain governance} for protocol tuning, reputation, and dispute resolution.
}


\subsection{Discussion}
One core challenge in decentralized training is the free-riding problem~\cite{Olson1965Logic}. To address this, a \PoL mechanism~\cite{Jia2021ProofOfLearning, thudi2022necessity}, requires participants to demonstrate that their model has been genuinely trained on the provided dataset, may be implemented. Techniques such as periodic accuracy or loss validation and comparison with expected learning curves could help in verifying that a model has been authentically trained~\cite{choi2024tools,srivastava2024optimistic}. Additionally, cryptographic methods such as \ZKPs~\cite{garg2023experimenting,abbaszadeh2024zero} could provide an additional layer of verification without exposing the model details, ensuring that participants have genuinely completed training tasks as specified by the protocol.

\gapbox{How to combine \PoL consensus, \ZKPs, or reputation scoring schemes with staking mechanisms to enhance the existing \DeAI training platforms?}
\vspace{-0.6\baselineskip}
\section{Post-Training}
\subsection{Inference}
\textbf{Model inference} applies the converged parameters $\theta^* \in \Theta$ to unseen inputs $x$ to generate predictions $y$. In \CeAI, inference is hosted on proprietary servers or clouds, where providers control access, efficiency, and pricing~\cite{LeCun2015DeepLearning}. 

\subsubsection{Challenges of Centralized Inference}
 \emph{Information Inefficiency}: outputs from $\theta^*$ are siloed at the central node, giving disproportionate advantages to the provider or select clients in domains such as finance, logistics, or governance. \emph{Inference Integrity}: with $\Gamma$ (governance) and $\Pi$ (incentives) fully controlled by the provider, users cannot verify whether the promised model was genuinely executed. This is critical in high-stakes domains such as medical diagnosis, or in commercial settings where providers may charge for premium models (e.g., o1-preview) while covertly serving downgraded substitutes such as GPT-3~\cite{LeCun2015DeepLearning,Sze2017EfficientProcessing}.

\subsubsection{Decentralized Solutions for Inference}
\DeAI protocols~\cite{numerai-docs, modulus-white-paper, hyperspace-white-paper, sertn-white-paper, ora-white-paper, allora-white-paper, rituals-docs} embed inference into $\mathcal{S}$ by coupling $G$ with validators $V_{\mathrm{val}}$ and incentives $\Pi$.

\paragraph{Incentivized Participation ($\Pi$)}
Allora~\cite{allora-white-paper} illustrates how incentives mitigate information inefficiency. Workers generate inferences and forecast losses, while reputers stake tokens to validate outputs via CometBFT consensus~\cite{CometBFT}. Rewards link to accuracy and stake, aligning incentives toward high-quality predictions.

\paragraph{Verification of Inference Integrity ($V_{\mathrm{val}}, \Gamma$)}
Two blockchain-based approaches enhance trust. (i) \ZKP-based inference~\cite{liu2021zkcnn, chen2024zkml, sun2024zkllm}, as in Sertn~\cite{sertn-white-paper}, providers prove correctness without exposing model or data, balancing privacy with verifiability. (ii) Optimistic proof-based inference, used in ORA~\cite{ora-white-paper}, assumes correctness by default but enables disputes through fraud-proof protocols~\cite{kalodner2018arbitrum}. Governance $\Gamma$ manages challenges and enforces accountability, while validators ensure results are auditable and tamper-resistant.

\insightbox{
Blockchain-backed model inference requires:
\ding{192} \emph{incentive-compatible payment mechanisms} to reward nodes for correct inference execution; and
\ding{193} \emph{verifiable inference schemes} to guarantee correctness of outputs in a trustless environment.
}

\subsubsection{Discussion}
The trade-off between \ZKP- and optimistic proof-based model inference hinges on security versus performance. The \ZKP-based approach ensures robust cryptographic protection for ML models but suffers from slower proof times as model size increases. In contrast, the optimistic proof-based approach leverages a fraud-proof system for model integrity, delivering better performance under specific trust assumptions. The choice between the two approaches should be determined by the specific requirements of the application scenario.

\gapbox{How to balance the trade-off between security and efficiency in \AI model inference?}
\vspace{-0.6\baselineskip}
\subsection{\AI Agents}

An autonomous system that can sense its environment, make decisions, and act toward goals is called an \emph{\AI agent}~\cite{poole2010artificial,castelfranchi1998modelling}. In practice, agents often embed trained models $\theta \in \Theta$ to guide behavior, while interacting with peers over $G$ and being governed by $\Gamma$ and incentivized by $\Pi$.

\subsubsection{Challenges in \CeAI Agents}
Centralized agent frameworks face three recurring challenges:
\ding{192} \emph{Scalability.} When $G$ reduces to a star topology, a central hub must coordinate all agent interactions. As the number of agents or tasks grows, this hub becomes a bottleneck for communication and compute allocation~\cite{montes2019distributed}.
\ding{193} \emph{Interoperability.} Isolated agent silos, each tied to proprietary infrastructures, limit cross-agent collaboration, preventing formation of heterogeneous agent collectives.
\ding{194} \emph{Trust and Security.} With both $\Gamma$ and $\Pi$ concentrated in a single provider, agents depend on central servers for decision-making and data storage, exposing them to tampering, single points of failure, or misaligned incentives.

\subsubsection{Solutions for \DeAI Agents}

Blockchain can address the challenges faced by \CeAI agents through many perspectives, as shown in industry protocols~\cite{fetch-ai-white-paper, arbius-white-paper, theoriq-white-paper, elna-white-paper, delysium-white-paper, open-serv-white-paper, autonolas-white-paper, open-agents-docs, shinkai-white-paper, balance-dao-docs}.

\noindent\textbf{Decentralizing Operating Environment.}  
Over $G$, agents interact without a central coordinator. Fetch.AI~\cite{fetch-ai-white-paper} demonstrates this by enabling autonomous economic agents to transact in an Open Economic Framework, while Delysium~\cite{delysium-white-paper} supports decentralized agent networks through layered communication, governance, and auditing. Theoriq~\cite{theoriq-white-paper} extends this with modular interoperability, allowing agents to form a dynamic mechanism to aggregate $\theta_i$ from different $D_i$.

\noindent\textbf{Enhancing Trust and Transparency.}
By encoding $\Gamma$ on-chain, agents can verify that interactions follow agreed-upon rules. For example, Morpheus~\cite{morpheus-white-paper} provides a decentralized cloud infrastructure where smart contracts enforce agreements, and all agent actions are immutably logged, reducing reliance on trusted intermediaries. 

\noindent\textbf{Incentivizing Agent Behavior.}
Through $\Pi$, tokenized incentive models reward agents for useful contributions of compute, data, or coordination. Fetch.AI leverages \uPoW to align incentives~\cite{fetch-ai-white-paper}, while Morpheus~\cite{morpheus-white-paper} rewards agents proportionally to their contributions. These schemes ensure that decentralized collectives of agents remain both economically sustainable and resistant to free-riding.
\vspace{-0.2\baselineskip}
\insightbox{
Blockchain can enhance \AI agents by:
\ding{192} providing a \emph{decentralized and tamper-resistant operating substrate} for agent actions and commitments;
\ding{193} enabling \emph{transparent, verifiable interactions} among agents via on-chain histories and smart contracts; and
\ding{194} supplying \emph{native incentive and reputation mechanisms} that reward cooperative behavior and penalize malicious actions.
}
\vspace{-0.2\baselineskip}
\subsubsection{Discussion}
Existing \DeAI agent protocols adopt various strategies to incentivize participation and maintain network integrity. As discussed, Fetch.AI~\cite{fetch-ai-white-paper} uses \uPoW protocol, where participants perform valuable computational tasks instead of traditional mining, encouraging broader participation and efficient resource use. In contrast, Theoriq~\cite{theoriq-white-paper} employs a combination of Proof-of-Contribution and Proof-of-Collaboration that focuses on reputation-based evaluation and collaborative optimization, with verifiable contributions strengthening trust. However, balancing computational contributions with reputation-based rewards remains a challenge, potentially introducing biases. Furthermore, scalable and robust reputation mechanisms are crucial to prevent manipulation and ensure network integrity as these systems grow.

\gapbox{How can we design scalable, robust, and unbiased incentive mechanisms to reward \DeAI agents fairly based on their contributions and collaborative efforts?}
\vspace{-0.3\baselineskip}
\subsection{Model Marketplaces}
\emph{\AI model marketplaces}~\cite{kumar2020marketplace} form the final stage of the \DeAI lifecycle. While the previous sections mostly discuss the supply side of the equation, \AI model marketplace is focused mainly on the demand for such \ML assets. Specifically, decentralized marketplaces operationalize $\Pi$ (incentives) and $\Gamma$ (governance) to distribute, monetize, and manage models. They enable end-users to browse, trade, and fine-tune trained models $\theta \in \Theta$, with provenance and ownership enforced on-chain.

\subsubsection{Challenges in Centralized Model Marketplaces}

Centralized marketplaces suffer from structural limitations.
\ding{192} \emph{Inequitable incentives:} Compensation schemes typically favor a few high-value or institutional models, leaving smaller contributors under-rewarded~\cite{Rodriguez2021Decentralizing}. This weakens long-term innovation.
\ding{193} \emph{Opaque rankings:} Marketplace visibility is often determined by proprietary algorithms~\cite{Andrews2022Algorithmic}, offering little transparency into how models are ranked or promoted. Users cannot verify whether $\Gamma$ (governance) reflects actual merit, reducing trust in the ecosystem.

\subsubsection{Blockchain-Based \DeAI Model Marketplaces}

\DeAI marketplaces integrate $\Pi$ and $\Gamma$ into open, verifiable systems in the following ways.

\noindent\textbf{Fair Incentive Mechanisms.} By tokenizing models as tradable digital assets, blockchain ensures provenance, secure ownership, and transparent compensation. BalanceDAO~\cite{balance-dao-docs} rewards contributors through token-based incentives combined with \ZKPs for integrity. SingularityNET~\cite{singularity-net-docs} similarly allows developers to directly publish and monetize models, reducing reliance on intermediaries. Sahara AI Marketplace~\cite{sahara-ai-docs} extends this approach by offering a decentralized hub for publishing, trading, and licensing both models and datasets.

\noindent\textbf{Transparent Model Ranking Algorithms.} 
With $\Gamma$ encoded on-chain, ranking and recommendation algorithms become auditable. For instance, Sahara AI~\cite{sahara-ai-docs} employs non-fungible receipts as verifiable ownership proofs, while reputation-based ranking ensures models are surfaced based on contribution quality rather than centralized bias.

\insightbox{Blockchain can improve \AI Model Marketplaces by providing: \ding{192} fair incentive mechanisms for model contributors; and \ding{193} transparent model ranking and recommendation algorithms.}

\subsubsection{Discussion}
The design of validation and provenance mechanisms is key to construct trust, transparency, and fair incentivization in \DeAI model marketplaces. Immutable Labs~\cite{immutable-labs-GPOW} and Sahara AI~\cite{sahara-ai-docs} provide unique but complementary answers to these issues. Immutable Labs focuses on comprehensive validation through a combination of trust scores, ML models, and validation controls to ensure high-quality contributions. On the other hand, Sahara AI Marketplace stresses asset provenance and verifies participants' identities, controls reputation, and guarantees AI ownership. While these approaches are effective individually, a comprehensive framework that integrates both robust validation and asset provenance remains underexplored, especially in addressing scalability and maintaining transparency without compromising model privacy.

\gapbox{How to design a scalable incentive mechanism that ensures robust validation, secure asset provenance, model privacy, and fair model exchange?}

\section{Security Risks and Mitigation Solutions}
\label{sec:security_risks}

In this section, to address \textbf{RQ4}, we analyze threats and mitigations (see Table~\ref{tab:deai-threats-scope}) and present empirical results on representative defenses across the \DeAI lifecycle. 
We further complement this analysis with a case study of Bittensor in Appendix~\ref{app:bittensor}.

\subsection{Threats to \DeAI}

\paragraph{Pre-Training}

\emph{Shared with \CeAI.}
In both centralized and decentralized contexts, data poisoning and backdoors remain major concerns: adversaries can craft samples that induce malicious behaviors. Recent works demonstrate their feasibility across modalities, including indiscriminate perturbations of feature extractors~\cite{Lu2024IndisPoison}, backdoors in medical vision-language models~\cite{Jin2024MedCLIP}, poisoning in graph prompt tuning~\cite{Song2025Krait}, and pruning-based poisoning in graph contrastive learning~\cite{Kato2024EdgePruner}. Similarly, \emph{privacy vulnerabilities} such as membership inference and reconstruction persist~\cite{Li2024ShakeToLeak,Ferry2024ProbReconstruct,Tao2025RaMIA}. Reliance on synthetic corpora or RAG pipelines exposes systems to \emph{task drift and derailment} through prompt injection~\cite{Abdelnabi2025TaskDrift}.

\emph{Unique to \DeAI.}
Compute infrastructure faces distinct risks in \DeAI: \emph{multi-tenant accelerator leakage} is exacerbated when GPUs are shared among untrusted nodes. Memory residues~\cite{Zhou2017VGMM,Sorensen2024LeftoverLocals}, microarchitectural side channels~\cite{Naghibijouybari2018RenderedInsecure}, cross-GPU channels~\cite{Dutta2022SpyGPUBox,Zhang2024BeyondBridge}, and uncore leaks even under MIG partitioning~\cite{Miao2024VeiledPathways} highlight the heightened risk surface introduced by adversarial co-location in \DeAI. Moreover, \emph{untrusted device execution} becomes more critical: while cloud providers can audit drivers and firmware, decentralized participants may run downgraded kernels to bypass GPU TEE protections~\cite{Volos2018Graviton,Ivanov2023SAGE,NVIDIA2024H100CC}. Finally, adversarial peers can weaponize \emph{fault injection and scheduler gaming}: undervolting silently corrupts computations~\cite{Qiu2021Lightning,Tsai2021NVBitFI}, while deadline-starvation attacks in shared clusters can bias training dynamics without detection.

\paragraph{On-Training}

\emph{Shared with \CeAI.}
A well-studied threat is \emph{Byzantine or poisoned updates}, where malicious participants submit manipulated gradients or adapters with the goal of corrupting the global model. Such updates range from indiscriminate perturbations that degrade overall performance to targeted manipulations that embed persistent backdoors. Prior work in \FL shows that even a single compromised client can implant high-accuracy triggers~\cite{Abad2023SniperFL,Bagdasaryan2020BackdoorFL,Blanchard2017Byzantine}. More recent studies demonstrate that collusion among multiple adversaries can further amplify their impact by overwhelming aggregation rules~\cite{Fang2020LocalModelPoison,Ngong2024OLYMPIA}.

\emph{Unique to DeAI.}
Collaborative training is especially vulnerable in \DeAI because updates come from many untrusted contributors, often incentivized by token rewards. This introduces both conventional FL risks and threats unique to decentralized, economically driven settings. Also, \emph{free-riding or model theft} occurs when nodes re-upload stolen or minimally altered adapters, unfairly claiming credit and undermining rewards~\cite{Jia2021ProofOfLearning}. Moreover, the lack of a central coordinator makes \emph{collusion under secure aggregation} damaging, as malicious groups can coordinate to mask poisoned updates or amplify their collective influence~\cite{BenItzhak2024ScionFL,Ngong2024OLYMPIA}. 

\paragraph{Post-Training}


\emph{Shared with \CeAI.}  
The final stage of the lifecycle involves deployment, sharing, and monetization of trained models. A key risk is \emph{model extraction}, in which adversaries can train surrogate models via black-box queries that closely approximate the original, thereby cloning proprietary assets and undermining contributor incentives~\cite{Karchmer2023ExtractionLimits,Juuti2019PRADA}. Another major challenge is \emph{prompt injection and task hijacking}, where malicious inputs alter model behavior or steer outputs away from intended goals~\cite{Abdelnabi2025TaskDrift}. Likewise, \emph{adversarial query abuse and jailbreaks} exploit vulnerabilities to bypass safety constraints~\cite{Debenedetti2024BreakingEggs,CodeLMSec2024}, threatening reliability in open-access ecosystems. In extreme cases, \emph{inference-time hijacking} enables attackers to subvert deployed models without retraining access~\cite{Ghorbel2018SnatchML}.  

\emph{Unique to \DeAI.}  
While these threats also affect \CeAI, their economic and trust implications are magnified in decentralized settings. Contributors in \DeAI rely on transparent attribution and reward mechanisms, making \emph{model extraction} especially damaging to incentives. Furthermore, the absence of a trusted central authority raises distinct challenges for \emph{inference integrity}: unlike in centralized systems, users cannot easily verify that service providers executed the promised model. This lack of verifiability underscores the need for cryptographic proofs of inference and on-chain lineage tracking in DeAI.

\begin{table*}[t]
\centering
\small
\caption{Security threats, representative mitigations, and scope across the \DeAI lifecycle}
\label{tab:deai-threats-scope}
\rowcolors{2}{rowgray}{white}
\renewcommand\arraystretch{1.05}
\begin{tabularx}{\textwidth}{>{\RaggedRight}p{2.7cm} Y Y >{\RaggedRight}p{1.0cm} >{\centering}p{0.8cm} >{\centering\arraybackslash}p{1.0cm}}
\toprule
\multirow{2}{*}{\textbf{Stage}} & \multirow{2}{*}{\textbf{Threat}} & \multirow{2}{*}{\textbf{Mitigation(s)}} & \multirow{2}{*}{\textbf{Type(s)}} & \multicolumn{2}{c}{\textbf{Scope}} \\
\cmidrule(lr){5-6}
& & & & \textbf{\CeAI} & \textbf{\DeAI} \\
\midrule

\rowcolor{stagegray}\multicolumn{6}{l}{\textbf{Pre-Training (Data)}}\\

& Data poisoning / backdoors\textsuperscript{\cite{Lu2024IndisPoison,Jin2024MedCLIP,Song2025Krait,Kato2024EdgePruner}} 
& Spectral signature filtering\textsuperscript{\cite{Tran2018Spectral}}; 
MixUp regularization\textsuperscript{\cite{Zhang2018MixUp}}; 
DP-SGD\textsuperscript{\cite{Abadi2016DPSGD}}
& \iconcrypto{} \icondetect{} \iconinfra{} 
& \blackcircle & \blackcircle \\

& Corrupted or unverifiable data provenance\textsuperscript{\cite{Kato2024EdgePruner}}
& Graph pruning\textsuperscript{\cite{Kato2024EdgePruner}}; 
dataset validation\textsuperscript{\cite{Gebru2018Datasheets}}
& \iconcrypto{} \iconinfra{} 
& \whitecircle & \blackcircle \\

& Membership inference / leakage\textsuperscript{\cite{Li2024ShakeToLeak,Ferry2024ProbReconstruct,Tao2025RaMIA}}
& Differential privacy\textsuperscript{\cite{Abadi2016DPSGD}}
& \iconcrypto{} 
& \blackcircle & \blackcircle \\

& Prompt-injected synthetic corpora (RAG drift)\textsuperscript{\cite{Abdelnabi2025TaskDrift}} 
& Data redaction\textsuperscript{\cite{Kong2024DataRedaction}}; 
unlearning\textsuperscript{\cite{Golatkar2020EternalSunshine}}; 
curated sources\textsuperscript{\cite{poisoncraft2025}}
& \icondetect{} 
& \whitecircle & \blackcircle \\

\rowcolor{stagegray}\multicolumn{6}{l}{\textbf{Pre-Training (Compute)}}\\

& Memory remanence / leakage\textsuperscript{\cite{Zhou2017VGMM,Sorensen2024LeftoverLocals}} 
& GPU buffer zeroing\textsuperscript{\cite{Zhou2017VGMM}}; 
driver scrubbing\textsuperscript{\cite{Sorensen2024LeftoverLocals}}
& \iconinfra{} 
& \blackcircle & \blackcircle \\

& GPU side channels (multi-tenant)\textsuperscript{\cite{Naghibijouybari2018RenderedInsecure,Dutta2022SpyGPUBox,Zhang2024BeyondBridge,Miao2024VeiledPathways}} 
& MIG partitioning\textsuperscript{\cite{NVIDIA2025MIGGuide}}; 
randomized scheduling\textsuperscript{\cite{Naghibijouybari2018RenderedInsecure}}; 
interconnect isolation\textsuperscript{\cite{Dutta2022SpyGPUBox,Zhang2024BeyondBridge}}
& \iconinfra{} 
& \blackcircle & \blackcircle \\

& Untrusted device execution\textsuperscript{\cite{Volos2018Graviton,Ivanov2023SAGE,NVIDIA2024H100CC}}
& GPU TEEs\textsuperscript{\cite{Volos2018Graviton,Ivanov2023SAGE,NVIDIA2024H100CC}}; 
remote attestation\textsuperscript{\cite{Sailer2004TPM}}
& \iconcrypto{} \iconinfra{} 
& \blackcircle & \blackcircle \\

& Fault injection / silent data corruption\textsuperscript{\cite{Qiu2021Lightning,Tsai2021NVBitFI}}
& ECC memory\textsuperscript{\cite{NVIDIA2010ECC}}; 
redundancy\textsuperscript{\cite{Reis2005SWIFT}}; 
runtime monitoring\textsuperscript{\cite{Tsai2021NVBitFI}}
& \icondetect{} \iconinfra{}
& \blackcircle & \blackcircle \\

& Co-location interference / scheduler gaming\textsuperscript{\cite{Miao2024VeiledPathways,Dutta2022SpyGPUBox,Zhang2024BeyondBridge}}
& Scheduling isolation\textsuperscript{\cite{Miao2024VeiledPathways}}; 
cluster governance\textsuperscript{\cite{Blaise2022ClusterSched}}
& \icongov{} \iconinfra{} 
& \whitecircle & \blackcircle \\

\rowcolor{stagegray}\multicolumn{6}{l}{\textbf{On-Training}}\\

& Byzantine / poisoned updates\textsuperscript{\cite{Abad2023SniperFL,Bagdasaryan2020BackdoorFL,Blanchard2017Byzantine,Fang2020LocalModelPoison,Ngong2024OLYMPIA}}
& Robust secure aggregation\textsuperscript{\cite{Bonawitz2017SecureAgg}}; 
anomaly detection\textsuperscript{\cite{Sun2019CanYou}}; 
slashing\textsuperscript{\cite{Jia2021ProofOfLearning}}
& \iconcrypto{} \iconecon{} \icondetect{} 
& \blackcircle & \blackcircle \\

& Free-riding / model stealing\textsuperscript{\cite{Jia2021ProofOfLearning}} 
& Watermarking\textsuperscript{\cite{Uchida2017DNNWM}}; 
commit--reveal\textsuperscript{\cite{Buterin2017PoS}}; 
\PoL\textsuperscript{\cite{Jia2021ProofOfLearning}}; 
ZKPs\textsuperscript{\cite{Groth2016Groth16}}
& \iconcrypto{} \iconecon{} \icondetect{} 
& \blackcircle & \blackcircle \\

& Collusion under secure aggregation\textsuperscript{\cite{BenItzhak2024ScionFL,Ngong2024OLYMPIA}}
& ZKPs\textsuperscript{\cite{Groth2016Groth16}}; 
cross-round auditing\textsuperscript{\cite{BenItzhak2024ScionFL}}
& \iconcrypto{} \icondetect{} 
& \whitecircle & \blackcircle \\

\rowcolor{stagegray}\multicolumn{6}{l}{\textbf{Post-Training}}\\

& Model extraction / cloning\textsuperscript{\cite{Karchmer2023ExtractionLimits,Juuti2019PRADA}} 
& Query-pattern detection\textsuperscript{\cite{Juuti2019PRADA}}; 
output perturbation\textsuperscript{\cite{Shejwalkar2021NoiseGuard}}; 
adaptive responses\textsuperscript{\cite{wang2016adversarial}}; 
output watermarking\textsuperscript{\cite{Kirchenbauer2023LLMWatermark}}
& \iconcrypto{} \icondetect{} 
& \blackcircle & \blackcircle \\

& Prompt injection / task hijacking\textsuperscript{\cite{Abdelnabi2025TaskDrift}}
& Activation-drift monitoring\textsuperscript{\cite{Abdelnabi2025TaskDrift}}; 
query sanitization\textsuperscript{\cite{PromptInject2022}}
& \icondetect{} 
& \blackcircle & \blackcircle \\

& Adversarial query abuse / jailbreaks\textsuperscript{\cite{Debenedetti2024BreakingEggs,CodeLMSec2024}} 
& Rate limiting\textsuperscript{\cite{OpenAI2023RateLimit}}; 
randomized output shaping\textsuperscript{\cite{Carlini2023SafetyShaping}}; 
safety filters\textsuperscript{\cite{Bai2022ConstitutionalAI}}
& \icondetect{} \iconinfra{} 
& \blackcircle & \blackcircle \\

& Inference-time model hijacking\textsuperscript{\cite{Ghorbel2018SnatchML}} 
& Output watermarking\textsuperscript{\cite{Uchida2017DNNWM,Kirchenbauer2023LLMWatermark}}; 
versioning + lineage tracking\textsuperscript{\cite{Ding2024ModelVersioning}}; 
unlearning\textsuperscript{\cite{Golatkar2020EternalSunshine}}
& \iconcrypto{} \icongov{} 
& \blackcircle & \blackcircle \\

\bottomrule
\end{tabularx}

\vspace{0.4em}
\renewcommand\arraystretch{1}
\begin{tabular}{ll@{\hspace{1.5em}}ll@{\hspace{1.5em}}ll}
\iconcrypto{} & Cryptographic defense &
\iconecon{}    & Economic / incentive mechanism &
\icondetect{}  & Detection / monitoring \\
\icongov{}     & Governance / community oversight &
\iconinfra{}   & System / infrastructure hardening &
\end{tabular}
\end{table*}

\subsection{Possible Mitigation Solutions}

\paragraph{Pre-training}

Pre-training aggregates heterogeneous, crowd-sourced data, so defenses must emphasize provenance, screening, and repair. Model-level defenses such as REStore~\cite{LeRoux2024REStore} provide black-box backdoor detection via rare-event probing, while domain-specific sanitization can filter poisoned samples~\cite{Kato2024EdgePruner}. Privacy-driven poisoning risks can be reduced through membership-inference protections~\cite{Li2024ShakeToLeak} and post-hoc redaction tools~\cite{Kong2024DataRedaction}, enabling adversarial samples to be removed even after ingestion.

On the compute side, GPU-level defenses address memory leakage, side channels, and untrusted execution. Memory remanence is mitigated through buffer zeroing~\cite{Zhou2017VGMM} and driver-level scrubbing~\cite{Sorensen2024LeftoverLocals}. Side-channel defenses include partitioning and noise~\cite{Naghibijouybari2018RenderedInsecure}, interconnect isolation~\cite{Dutta2022SpyGPUBox}, NVLink scheduling~\cite{Zhang2024BeyondBridge}, and uncore monitoring~\cite{Miao2024VeiledPathways}. \TEEs such as Graviton~\cite{Volos2018Graviton}, SAGE~\cite{Ivanov2023SAGE}, and NVIDIA H100~\cite{NVIDIA2024H100CC} enable attested, isolated execution. For fault injection and corruption, NVBitFI~\cite{Tsai2021NVBitFI} supports resilience testing, while ECC memory~\cite{NVIDIA2010ECC} and redundancy offer practical mitigation. At the cluster level, strict MIG partitioning~\cite{NVIDIA2025MIGGuide} provides isolation, though adversary-aware QoS remains necessary~\cite{Miao2024VeiledPathways}.

\paragraph{On-Training}
Defenses during decentralized training focus on preventing malicious clients from subverting model updates. \emph{Robust and quantized secure aggregation} restricts adversarial influence, with protocols like ScionFL~\cite{BenItzhak2024ScionFL} and OLYMPIA~\cite{Ngong2024OLYMPIA} combining efficiency with Byzantine robustness for large-scale use. \emph{\PoL}~\cite{Jia2021ProofOfLearning}, \emph{cross-round anomaly detection}, \emph{economic penalties} or \emph{cryptographic guarantees} like Groth16~\cite{Groth2016Groth16} can also be helpful. A parallel risk is \emph{model stealing and free-riding}, where participants claim credit for unearned updates. Mitigations include: (i) \emph{watermarking and fingerprinting} to embed detectable signatures; (ii) \emph{commit–and–reveal protocols} to enforce provenance; (iii) \emph{\PoL} logs to attest training legitimacy; and (iv) \emph{\ZKPs} to validate updates without disclosing details. Combined, these mechanisms deter free-riding and preserve the fairness of decentralized incentives.

\paragraph{Post-Training}
Once models are deployed, mitigating unauthorized replication requires both detection and deterrence. \emph{PRADA}~\cite{Juuti2019PRADA} detects anomalous query patterns linked to extraction, enabling operators to throttle or block adversaries. \emph{Watermarking} of adapters or outputs~\cite{Uchida2017DNNWM,Kirchenbauer2023LLMWatermark} embeds verifiable signatures so stolen models can be attributed to their source. \emph{Model versioning and lineage tracking}~\cite{Ding2024ModelVersioning} links deployed models to specific contributors, supporting accountability and forensics. Finally, \emph{activation-drift monitoring}~\cite{Abdelnabi2025TaskDrift} detects abnormal behavior caused by prompt injection or task hijacking, enabling early intervention before failures propagate across the ecosystem. 

\subsection{Evaluation Experiments}
To complement our taxonomy and threat analysis, we empirically evaluate representative mitigation strategies across the \DeAI lifecycle. Specifically, drawing on Table~\ref{tab:deai-threats-scope}, we select one high-impact threat from each stage—\emph{data poisoning} in pre-training, \emph{model stealing} in on-training, and \emph{model extraction} in post-training—and benchmark multiple defenses for each. All experiments were conducted on a MacBook Pro with an Apple M2 Max CPU and 96 GB unified memory.

\subsubsection{Pre-Training}

\paragraph{Experimental Setup}
We study \DeAI data poisoning attacks~\cite{Lu2024IndisPoison,Jin2024MedCLIP,Song2025Krait,Kato2024EdgePruner}, where adversaries inject backdoor samples that maintain clean accuracy but enforce hidden behaviors. We evaluate five defenses: \textit{Baseline Training}; \textit{Spectral Signature Filtering}~\cite{Kato2024EdgePruner} for clustered-sample detection; \textit{DP-SGD}~\cite{Li2024ShakeToLeak} to limit single-sample influence; \textit{MixUp}~\cite{Song2025Krait} for decision-boundary smoothing; and a combined \textit{Spectral + DP-SGD} pipeline. The experiments employ a compact CNN with GroupNorm trained on Fashion-MNIST under poison rates of 0.5\%, 1\%, 2\%, and 5\%. Backdoors are $3{\times}3$ pixel triggers targeting the “sneaker’’ class. Models train with AdamW for 10 epochs, with three seeds per condition (60 runs total). We report \textit{Attack Success Rate} (ASR), \textit{Clean Accuracy} (CA), and overhead metrics (filtered data fraction, training time, privacy budget). This setup quantifies robustness–utility trade-offs in decentralized pre-training. Full details appear in Appendix~\ref{app:pre_train_setup}.

\paragraph{Results}
Table~\ref{tab:poison_results} shows clear trade-offs across defenses and poison rates. The baseline maintains high clean accuracy ($\sim$91\%) but exhibits near-perfect attack success (ASR $\geq$98\%) throughout. DP-SGD is the most robust at low poisoning (ASR 51.3\% at 0.5\%) but degrades as poisoning increases (98.0\% at 5\%) and consistently incurs heavy utility costs (–7–8\% accuracy, $\sim$3$\times$ runtime). MixUp preserves accuracy ($\sim$92\%) and lowers runtime but provides inconsistent robustness—helpful at 1\% poisoning (92.0\%) yet ineffective at higher rates (99.9–100\%). Spectral Filtering removes only 1\% of data and leaves ASR essentially unchanged. The combined approach (Spectral + DP-SGD) reduces ASR relative to Spectral alone (e.g., 87.0\% vs.\ 95.8\% at 0.5\%) and improves utility over DP-SGD, but remains less robust and offers no efficiency gains. Overall, DP-SGD is preferable at low poisoning, MixUp offers good utility but is brittle under stronger attacks, and complex filtering adds limited benefit. These results suggest that lightweight regularization (e.g., MixUp) may be more practical than heavy filtering for decentralized pre-training. Full results appear in Appendix~\ref{sec:app:pre_train_results}.

\begin{table}[htbp]
\centering
\caption{Simulation of backdoor data poisoning mitigation}
\label{tab:poison_results}
\renewcommand\arraystretch{0.9}
\resizebox{\linewidth}{!}{
\begin{tabular}{lcccccc@{}}
\toprule
\textbf{Mitigation} & \textbf{Poison Rate} & \textbf{ASR} & \textbf{Clean Acc.} & \textbf{Removed\%} & \textbf{Time (s)} \\
\midrule

\multirow{4}{*}{Baseline} 
 & 0.5\% & 98.4 & 91.7 & 0.0 & 232 \\
 & 1.0\% & 98.3 & 91.7 & 0.0 & 258 \\
 & 2.0\% & 97.9 & 91.3 & 0.0 & 232 \\
 & 5.0\% & 99.8 & 91.2 & 0.0 & 228 \\

\midrule
\multirow{4}{*}{Spectral\textsuperscript{\cite{Tran2018Spectral}}} 
 & 0.5\% & 95.8 & 91.4 & 1.0 & 232 \\
 & 1.0\% & 99.3 & 91.4 & 1.0 & 259 \\
 & 2.0\% & 99.1 & 91.3 & 1.0 & 246 \\
 & 5.0\% & 99.9 & 91.8 & 1.0 & 241 \\

\midrule
\multirow{4}{*}{DP-SGD\textsuperscript{\cite{Abadi2016DPSGD}}}
 & 0.5\% & 51.3 & 84.5 & 0.0 & 787 \\
 & 1.0\% & 83.4 & 83.7 & 0.0 & 764 \\
 & 2.0\% & 90.4 & 84.6 & 0.0 & 784 \\
 & 5.0\% & 98.0 & 83.4 & 0.0 & 783 \\

\midrule
\multirow{4}{*}{MixUp\textsuperscript{\cite{Zhang2018MixUp}}}
 & 0.5\% & 98.1 & 92.0 & 0.0 & 231 \\
 & 1.0\% & 92.0 & 92.0 & 0.0 & 204 \\
 & 2.0\% & 99.9 & 92.2 & 0.0 & 249 \\
 & 5.0\% & 100.0 & 92.1 & 0.0 & 237 \\

\midrule
\multirow{2}{*}{Spectral\textsuperscript{\cite{Tran2018Spectral}}}
 & 0.5\% & 87.0 & 88.3 & 1.0 & 734 \\
 & 1.0\% & 96.4 & 87.8 & 1.0 & 761 \\
\multirow{2}{*}{+ DP-SGD\textsuperscript{\cite{Abadi2016DPSGD}}}
 & 2.0\% & 97.0 & 87.8 & 1.0 & 775 \\
 & 5.0\% & 98.9 & 87.9 & 1.0 & 760 \\

\bottomrule
\end{tabular}
}
\begin{tablenotes}
\footnotesize
\item \textbf{ASR}: Attack Success Rate. 
\item \textbf{Clean Acc.}: accuracy on unpoisoned test data.
\item \textbf{Removed\%}: fraction of training data filtered out.
\item \textbf{Time}: end-to-end training time per run.
\end{tablenotes}
\end{table}

\subsubsection{On-Training}

\paragraph{Experimental Setup}
We study model stealing during on-training in \DeAI platforms, where participants may submit pre-trained or stolen models instead of performing genuine training. We evaluate five defenses: \textit{Baseline Training} (honest vs.\ stolen LoRA-adapted model), \textit{Watermarking} (WM) for post-hoc identification, \textit{Commit-and-Reveal} (C–R) for cryptographic access control, \textit{\PoL} for verifiable training evidence, and \textit{\ZKPs} for property verification without revealing parameters. Each is assessed for theft-detection accuracy, performance on held-out data, compute overhead, storage cost, and verification time under matched conditions. We focus on \emph{adapter theft}, which reflects modern fine-tuning workflows. Using Qwen2.5-7B as the base model, we apply LoRA fine-tuning (rank=8, $\alpha$=16, 4 epochs, batch size=1 with gradient accumulation=8, learning rate $2{\times}10^{-4}$). LoRA adapters ($\approx$10MB) are far smaller than the full model ($\sim$14GB), making them realistic theft targets. The dataset includes a 200-sample training subset from \texttt{Hutao\_furina\_roleplay}~\cite{hutao_furina_roleplay}, two 10-sample fingerprint sets (animals, foods) for Watermarking, and a 1K held-out test set. All defenses are evaluated under identical theft scenarios. Full details are provided in Appendix~\ref{sec:app:on_train_setup}.

\begin{table}[htbp]
\centering
\caption{Comparison of model stealing mitigations}
\label{tab:model_stealing_results}
\renewcommand\arraystretch{1.}
\resizebox{\linewidth}{!}{
\begin{tabular}{lccrrr}
\toprule
\textbf{Mitigation} & \textbf{Eff.} & \textbf{Perf.} & \textbf{Comp. (s)} & \textbf{Storage (MB)} & \textbf{Ver. (s)} \\
\midrule

Baseline & No & 1.710 & N/A & N/A & N/A \\
\hline

WM\textsuperscript{\cite{Uchida2017DNNWM}} 
 & Yes & 1.924 & 893 & 0.005 & \textbf{$<$0.001} \\

C-R\textsuperscript{\cite{Buterin2017PoS}} 
 & Yes & 1.710 & 6,247 & 12.9 & 0.050 \\

PoL\textsuperscript{\cite{Jia2021ProofOfLearning}} 
 & Yes & \textbf{1.697} & 4,498 & 0.007 & 0.010 \\

ZKP\textsuperscript{\cite{Groth2016Groth16}} 
 & Yes & 1.710 & \textbf{0.18} & \textbf{0.0006} & 0.092 \\

\bottomrule
\end{tabular}
}
\begin{tablenotes}
\footnotesize
\item Eff.: binary theft-detection outcome (Yes = perfect, No = none).
\item Perf.: model loss on the evaluation set (lower is better).
\item Comp.: computational cost for end-to-end protection runtime (seconds).
\item Storage: additional storage overhead for protection metadata (MB).
\item Ver.: time to verify theft (seconds).
\item Bold indicates the best value per metric among methods.
\end{tablenotes}
\end{table}

\paragraph{Results}
Our evaluation (Table~\ref{tab:model_stealing_results}) shows that all four defenses—Watermarking, C–R, PoL, and \ZKPs—achieve perfect theft detection (100\%), whereas the baseline provides none (Perf.\ loss = 1.710). For model utility, PoL performs best (1.697), slightly improving on the baseline; C–R and ZKP match baseline utility (1.710), while Watermarking degrades performance most (1.924). These results indicate that strong theft detection is compatible with maintaining model quality.

Efficiency varies significantly. ZKP is by far the fastest (0.18s) and most storage-efficient (0.0006 MB), compared to WM (893s, 0.005 MB), PoL (4,498s, 0.007 MB), and C–R (6,247s, 12.9 MB). Verification times show a similar pattern: WM $\le$ 0.001s, PoL 0.010s, C–R 0.050s, and ZKP 0.092s. In summary, ZKP offers the best overall efficiency–privacy profile, PoL achieves the strongest utility, Watermarking is simple but costly in accuracy, and C–R provides temporal protection with high overhead. Full details are reported in Appendix~\ref{app:on_train_results}.

\subsubsection{Post-Training}

\paragraph{Experimental Setup}
Model extraction attacks exploit post-deployment API access: adversaries query a model and train surrogates that mimic its behavior, threatening usage-based rewards and intellectual property in \DeAI systems. We evaluate five defenses: \textit{Query Pattern Detection} (sequence monitoring), \textit{Output Watermarking} (forensic signatures), \textit{Rate Limiting} (per-client quotas), \textit{Output Perturbation} (noise injection), and \textit{Adaptive Responses} (diversified outputs). Like in the pre-training study, we deploy a pre-trained CNN on Fashion-MNIST behind a simulated inference API. Attackers issue 500 crafted queries (test inputs with added noise) and train surrogate models of the same architecture for 20 epochs. We measure (i) extraction success via surrogate accuracy relative to the target, and (ii) computational overhead via extraction time, with each condition run three times (18 runs total). This setup exposes the security–utility trade-offs of post-training defenses and informs which mechanisms best limit extraction while maintaining practical inference performance. Full details appear in Appendix~\ref{sec:app:post_train_setup}.

\begin{table}[t]
\centering
\caption{Model Extraction Defense Evaluation Results}
\label{tab:defense_results}
\renewcommand\arraystretch{1.0}
\resizebox{\linewidth}{!}{
\begin{tabular}{lrr}
\toprule
\textbf{Mitigation} & \textbf{Surrogate Accuracy} & \textbf{Extraction Time (s)} \\
\midrule
Baseline & 15.32\%  & 8.60 \\
\hline

Query Pattern Detection\textsuperscript{\cite{Juuti2019PRADA}} 
 & 15.84\% & 30.96 \\

Output Watermarking\textsuperscript{\cite{Kirchenbauer2023LLMWatermark}} 
 & 14.88\% & 8.78 \\

Rate Limiting\textsuperscript{\cite{OpenAI2023RateLimit}} 
 & 20.35\% & 8.73 \\

Output Perturbation\textsuperscript{\cite{Shejwalkar2021NoiseGuard}} 
 & \textbf{13.29\%} & \textbf{8.62} \\

Adaptive Responses\textsuperscript{\cite{wang2016adversarial}} 
 & 16.30\% & 8.65 \\

\bottomrule
\end{tabular}
}
\begin{tablenotes}
\small
\item Results are averaged across three independent runs.
\item Surrogate Accuracy: lower is better.
\item Target model accuracy on the held-out set (no defenses): 84.89\%.
\end{tablenotes}
\end{table}

\paragraph{Results}
Table~\ref{tab:defense_results} summarizes the effectiveness of five extraction defenses, averaged over three runs. Surrogate accuracies range from 13.29\% to 20.35\%, indicating substantial variation in protection strength. Output Perturbation is most effective (13.29\%), reducing accuracy by 2.03 points relative to the baseline (15.32\%) while keeping extraction time low (8.62s). Rate Limiting performs worst (20.35\%), showing that throttling alone does not stop extraction when queries can be distributed over time. Output Watermarking and Adaptive Responses provide moderate protection (14.88\% and 16.30\%), whereas Query Pattern Detection offers minimal accuracy benefit (15.84\%) but incurs heavy overhead (30.96s). Overall, access-based defenses (Rate Limiting, Query Pattern Detection) provide limited robustness, either allowing effective extraction or imposing high runtime costs. In contrast, response-based methods that degrade the adversary’s learning signal—especially Output Perturbation—consistently reduce surrogate fidelity at minimal cost. These findings suggest that degrading output quality is more effective than restricting access for post-training protection in \DeAI systems.

\subsubsection{Discussion}
Across all three stages, our experiments reveal consistent patterns that map directly onto the \DeAI taxonomy (Sec.~\ref{sec:deai-training}) and threat matrix (Table~\ref{tab:deai-threats-scope}). In pre-training, lightweight defenses such as MixUp offer good utility but fail under stronger poisoning, while stronger methods like DP-SGD are costly and only partly effective—pointing to a need for \emph{robustness techniques suited to decentralized settings}. In on-training, protocol-layer cryptographic tools (PoL, ZKPs, commit-and-reveal) reliably detect theft but differ sharply in computation, storage, and verification costs, clarifying which are feasible for resource-constrained networks. In post-training, access-based controls (rate limiting, query pattern detection) offer weak protection, whereas response-based defenses (output perturbation, watermarking) more effectively degrade surrogate quality. Collectively, these results anchor our taxonomy in concrete trade-offs and motivate the research gaps detailed in Sec.~\ref{sec:security_risks} and Appendix~\ref{app:open-research-challenges}.

\section{Conclusion}

In this \SoK, we systematize the emerging landscape of blockchain-enabled \DeAI. We formalize \DeAI as a framework that spans across different stages, and propose a taxonomy that situates existing protocols across this lifecycle. We show how \DeAI can address core limitations of \CeAI. Our threat analysis also reveals that \DeAI inherits many traditional \CeAI risks while introducing new attack surfaces. We review representative mitigations and report experimental evidence of their effectiveness, highlighting both progress and open challenges. Overall, this work provides a foundation for understanding \DeAI as both an opportunity and a risk. We hope our taxonomy, security evaluation, and identification of research gaps will guide future efforts toward building secure, trustworthy, and sustainable \DeAI systems.

\section*{LLM Usage Considerations} LLMs were used for editorial purposes in this manuscript, and all outputs were inspected by the authors to ensure accuracy and originality.

\section*{Acknowledgments}
This work was partially supported by a research grant from the Ethereum Foundation.



\bibliography{references}
\bibliographystyle{IEEEtran}

\appendix

\begin{figure*}[t]
\centering
\includegraphics[width=\linewidth]{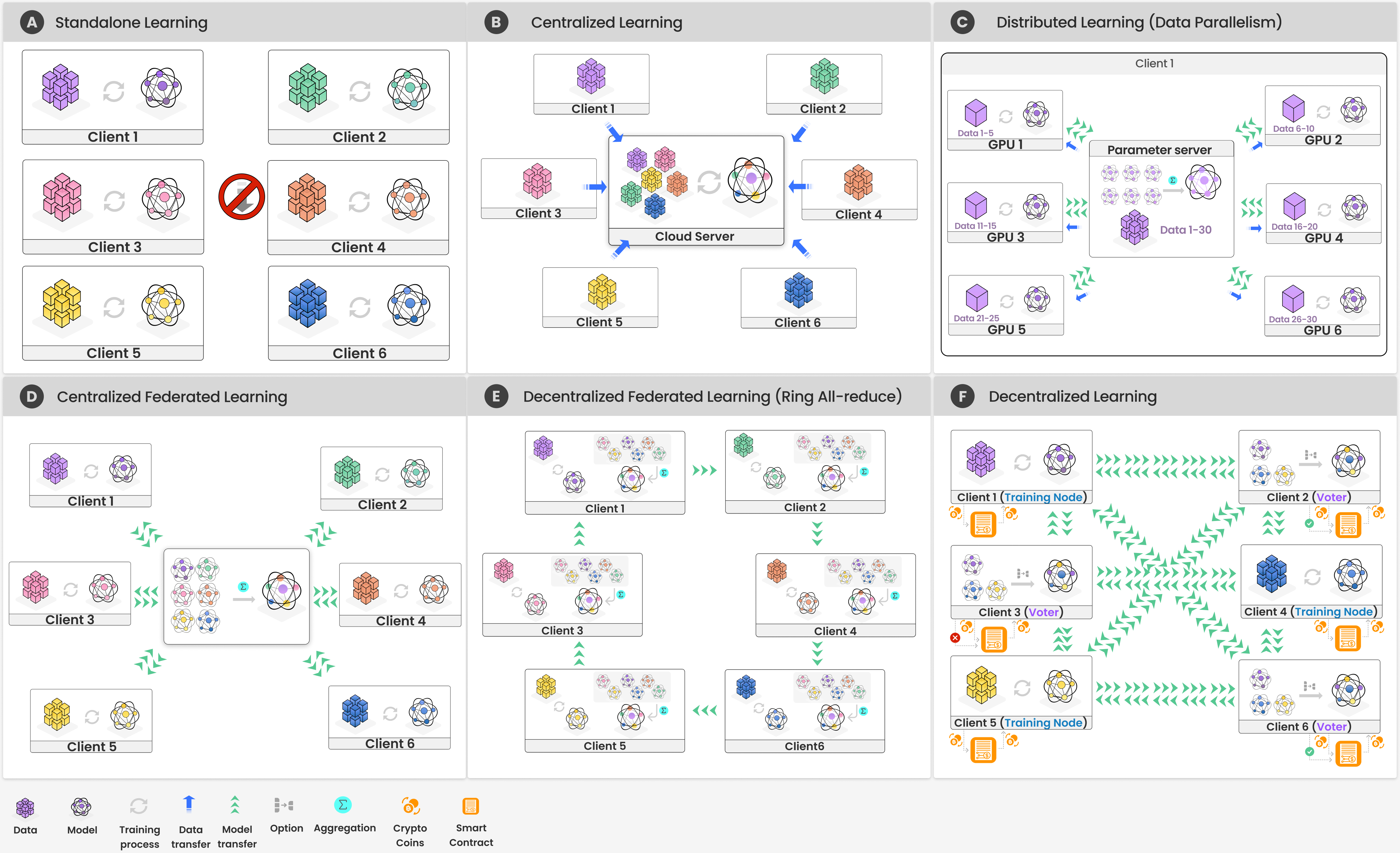}
\caption{Comparison of different \ML paradigms: (A) Standalone Learning, (B) Centralized Learning, (C) Distributed Learning (Data Parallelism), (D) Centralized \FL, (E) Decentralized Federated Learning (Ring All-reduce), and (F) Decentralized Learning.}
\label{fig:learning_para_stru}
\end{figure*}

\subsection{Supporting Figures}\label{app:additional-figures}
\begin{figure}[htbp]
    \centerline{\includegraphics[width=1\linewidth]{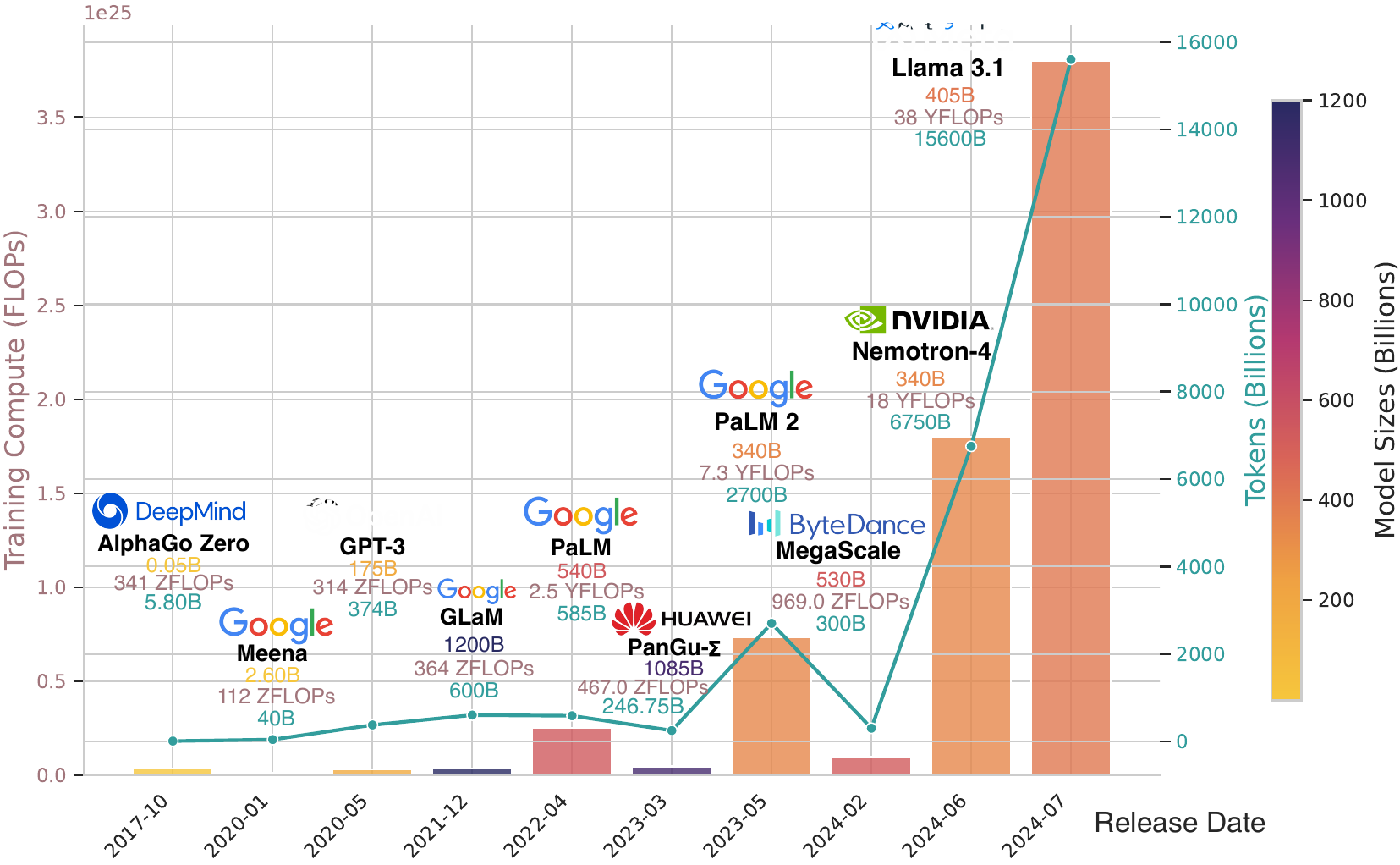}}
    \caption{Trends in the development of large-scale AI models, showcasing Training Compute Costs (bar chart), Tokens (or Data points) Counts (line plot), and Model Sizes (heatmap). Data sourced directly from~\cite{epochAIModels}.}
    \label{fig:compute_model_size_trend}
\end{figure}

Figure~\ref{fig:compute_model_size_trend} illustrates the accelerating growth of large-scale AI models, where computational demands, model size, and training data volumes are all rising sharply. These escalating requirements strain centralized infrastructures, motivating the exploration of decentralized approaches for provisioning compute resources.

\subsection{Related Work}
\label{app:related-work}

{Research at the intersection of decentralization, learning, and blockchain spans multiple communities, and several surveys have attempted to systematize different parts of this landscape. We group and analyze these efforts to clarify how our work complements and differs from prior surveys.

\subsubsection{Decentralized Learning without Blockchain}

While the scope of this paper falls squarely in the remit of blockchain-based \DeAI solutions, we reckon \DeAI is a broad research area that extends far beyond blockchain. A substantial body of work explores decentralized and peer-to-peer learning without relying on distributed ledgers. Classical methods such as D-PSGD~\cite{lian2017decentralized} and gossip-based learning~\cite{hegedus2019gossip} provide communication-efficient and serverless learning under benign settings. Beyond these works, recent advances in \emph{Byzantine-robust decentralized learning} address adversarial model updates and data heterogeneity. Examples include mixing-based robustness~\cite{Allouah2023Mixing}, bucketing-based aggregation~\cite{Karimireddy2021Bucketing}, tight robustness bounds under heterogeneity~\cite{Allouah2023Heterogeneous}, and variance-reduction-based defenses~\cite{Gorbunov2023Variance}. However, despite their algorithmic strengths, these non-blockchain decentralized methods generally lack built-in primitives for \emph{trust}, \emph{incentive alignment}, and \emph{verifiable coordination} among mutually distrustful participants. This gap motivates growing interest in blockchain-backed \DeAI, where blockchain and smart contracts can furnish tamper-resistant execution, transparent auditability, and automated incentive mechanisms. Consequently, while \DeAI is much richer than blockchain-based approaches alone, this survey focuses specifically on blockchain-enabled \DeAI due to its unique capabilities in addressing trust and incentive challenges in open, adversarial environments.

\subsubsection{Decentralized Federated Learning}

A parallel research line—recently surveyed in~\cite{Ehsan2024DFLsurvey,Liangqi2024DFLsurvey,Martinez2023DFLsurvey,Gabrielli2023DFLsurvey,Leon2022IncentivizedFL}—examines \emph{decentralized federated learning} (DFL), in which model aggregation is performed through peer-to-peer protocols rather than a central server. These works provide taxonomies over communication topologies, aggregation strategies, privacy mechanisms, and protocol-level challenges in decentralized FL. However, DFL surveys generally assume cooperative or partially adversarial environments and do not address open, permissionless settings where nodes may be economically motivated, sybil-prone, or rationally malicious. Moreover, they do not analyze blockchain as a substrate for coordination, incentive engineering, or verifiable execution. As such, their threat models, design axes, and system assumptions differ substantially from blockchain-based \DeAI.

\subsubsection{Blockchain and Machine Learning}

Several existing surveys explore how blockchain can enhance aspects of \ML. Kayikci and Khoshgoftaar review cross-domain applications of blockchain for \ML, covering their integrated use in finance, medicine, supply chain, and security~\cite{Kayikci2024BlockchainML}. However, their survey largely treats the ML pipeline as a homogeneous, static construct and does not provide a systematic understanding of how blockchain interacts with specific stages of the learning lifecycle (e.g., data preparation, training, inference, evaluation). Moreover, although the survey briefly references real-world examples such as the DHL Global Trade Barometer~\cite{DHLTradeBarometer}, the majority of its analysis focuses on conceptual or academic works, offering limited empirical insight into how blockchain-enabled ML systems operate in practice. Ural and Yoshigoe, on the other hand, focus on “blockchain-enhanced ML”, highlighting how blockchain can serve as a secure and immutable platform for data sharing, model validation, and task execution~\cite{Ural2023BlockchainEnhancedML}. While their survey offers a useful overview of scholarly work at the intersection of ML and blockchain---and provides a notable examination of emerging ideas such as uPoW---it devotes comparatively little attention to the crypto-native incentive structures that underpin blockchain-enabled ML ecosystems. These incentive mechanisms are central to ensuring that decentralized ML systems remain economically sustainable, self-sufficient, and resilient in open, permissionless environments, and therefore constitute a core aspect of our analysis.
Earlier works provide broader reviews of blockchain-for-AI integration~\cite{Khaled2019BlockchainAI,Adedoyin2021BlockchainAI}, typically emphasizing high-level opportunities and application stories, but they share similar limitations in that they neither characterize the full lifecycle of blockchain-enabled ML nor analyze the economic and operational dynamics of deployed systems.

Complementary lines of work systematize \emph{blockchain-based federated learning} (BFL)~\cite{Li2022BCFL,Wang2024BlockchainedFL,Ning2024BFLSurvey,Wu2023BFLSurvey,Cai2024BCFL}. These surveys examine how blockchain can replace or augment the federated learning aggregator, focusing on on-chain aggregation, reputation scoring, and secure update logging.

While collectively valuable, these surveys share several limitations from the perspective of blockchain-based \DeAI.  
(1) Most treat blockchain as an \emph{auxiliary infrastructure} for ML or FL, not as the fundamental coordination substrate for open, permissionless AI ecosystems.  
(2) They devote limited attention to blockchain-native concepts essential for \DeAI, such as tokenomics, incentive compatibility, sybil resistance, economic consensus, and on-chain governance.  
(3) BFL surveys remain scoped to federated learning and do not encompass broader decentralized AI paradigms such as open model marketplaces, decentralized inference networks, agent-based systems, or protocol-level curation and reputation.  
(4) Existing surveys primarily offer conceptual taxonomies and lack empirical analysis of deployed systems (e.g., gas costs, throughput, adversarial robustness under economic incentives).  
(5) Most predate the recent emergence of incentive-driven \DeAI platforms (e.g., Bittensor\cite{BittensorWhitepaper}, FLock\cite{flock-white-paper}), leading to incomplete coverage of current design patterns and practical considerations.

Against this backdrop, our SoK differs from prior surveys in three important ways. First, we treat \emph{blockchain-based \DeAI as a distinct system class}, not merely as ML augmented with blockchain components. We develop a cross-layer taxonomy that jointly considers the learning pipeline, blockchain infrastructure, incentive and governance mechanisms, verification models, and deployment architectures.  
Second, we analyze both academic proposals and real-world deployed \DeAI platforms, providing empirical measurements of their operational costs, performance characteristics, and robustness properties.  
Third, we develop a unified threat and trust model that captures adversaries unique to permissionless, economically motivated \DeAI ecosystems—attacks and incentives that prior surveys do not examine.  
Together, these contributions fill a gap in the literature by offering the first systematic, empirically grounded, and blockchain-centered analysis of fully decentralized AI ecosystems.

\subsection{Task Proposing} \label{app:task-proposing}
This stage marks the beginning of a \DeAI model's lifecycle, initiated by the demand of a task creator who specifies the objectives, data requirements, and evaluation criteria for downstream training. Unlike in \CeAI, where tasks are defined internally by a central authority, task proposing in \DeAI must occur in an open environment with heterogeneous and potentially untrusted participants. This shift introduces unique challenges: proposals must be validated for legitimacy, feasibility, and fairness without relying on a single trusted coordinator. In our formulation, task proposing explicitly encompasses two preparatory pillars, \emph{Distributed Learning Algorithm Preparation} and \emph{Code Verification}, that precede training.

We place this discussion in the Appendix because, to date, none of the surveyed \DeAI protocols provide fully decentralized task proposing mechanisms. Current systems typically assume a centralized initiator, leaving open research questions on how to design secure, verifiable, and incentive-compatible task markets. Accordingly, we structure this appendix to first present \emph{Distributed Learning Algorithm Preparation} and \emph{Code Verification}, and then analyze \emph{Security Threats of Task-Proposing}, followed by \emph{Mitigation Solutions of Task-Proposing}. This organization lays a foundation for an underexplored yet critical stage of the lifecycle by connecting concrete preparatory steps with their attendant risks and defenses.

\subsubsection{Distributed Learning Algorithm Preparation}

In DeAI systems, selecting and designing appropriate learning algorithms is essential for ensuring data privacy and communication efficiency. In zero-trust environments, where participants lack inherent trust in one another, the algorithms must satisfy key requirements: \ding{192} enable efficient learning with minimal information exchange, \ding{193} maximize data privacy by avoiding direct transfers of sensitive or any raw data, and \ding{194} reduce communication overhead through limited and efficient message exchanges. Existing distributed learning approaches can be generally classified into data-sharing, model-sharing, knowledge-sharing, and result-sharing methodologies~\cite{ma2023trusted, le2024applications}.

\noindent
\textbf{Data-sharing.} In this approach, DeAI systems centralize private or anonymized data for aggregation and training, enabling robust learning outcomes. However, techniques such as multi-agent reinforcement learning (MARL)~\cite{littman1994markov} raise concerns about information leakage (e.g., states, actions, rewards), computational overheads, and latency in large-scale applications. Solutions like QPLEX~\cite{wang2020qplex} and UPDeT~\cite{huupdet} enhance scalability but face bottlenecks in dynamic and resource-constrained environments.


\noindent
\textbf{Model-sharing.} This method emphasizes decentralized model updates, preserving privacy by transmitting model parameters instead of raw data. Synchronous techniques, such as Distributed SGD~\cite{shokri2015privacy}, mitigate computational delays but impose high communication costs. \FL~\cite{mcmahan2017communication} alleviates this by allowing multiple updates before aggregation, yet issues like gradient staleness in asynchronous approaches, such as Asynchronous Federated Learning (AFL)~\cite{xie2019asynchronous} warrant further optimization to balance efficiency and security.


\noindent
\textbf{Knowledge-sharing.} Leveraging knowledge distillation (KD)~\cite{hinton2015distilling} and split learning techniques (e.g., Splitfed~\cite{thapa2022splitfed}), this approach extracts insights from local datasets to inform global models. While privacy is maintained by keeping raw data local, challenges such as data heterogeneity and training complexity hinder scalability and generalization.


\noindent
\textbf{Result-sharing.} This strategy shares only the final outcomes, ensuring maximum privacy for sensitive domains such as healthcare. Methods like PATE-GAN~\cite{jordon2018pate} generate synthetic data to approximate true distributions, though inconsistencies in heterogeneous data environments remain problematic.



\paragraph{Challenges in Distributed Learning Algorithm Preparation} 

\noindent \textbf{Privacy and Security Challenges}
Privacy and security are primary concerns in decentralized learning. For example, in MARL, sensitive information shared among agents (e.g., states, actions, rewards) may lead to data leakage. Complex encryption techniques and differential privacy mechanisms protect privacy but are computationally expensive~\cite{shokri2015privacy}. In decentralized settings with low trust, asynchronous updates or delayed synchronization could allow attackers to compromise data integrity and model accuracy~\cite{kairouz2021advances}. 


\noindent \textbf{Scalability and Communication Efficiency}
As the number of participating nodes increases, communication efficiency becomes a major issue~\cite{mcmahan2017communication}. Frequent data sharing, model updates, and knowledge transfer across nodes can create high communication costs, and real-time coordination in dynamic environments can lead to latency. Techniques such as model partitioning and asynchronous updates improve communication efficiency but still face constraints in bandwidth and computational resources. Balancing efficient learning with minimal information exchange, especially in large-scale scenarios, remains a critical challenge.

\noindent \textbf{Model Consistency}
Ensuring consistency across models at different nodes is complex, particularly with asynchronous updates and heterogeneous data distributions. Independent updates can cause model drift and misalignment in learning objectives~\cite{li2020federated}. Approaches such as AFL allow asynchronous model updates but introduce delays, gradient staleness, and inconsistency, slowing convergence and reducing accuracy~\cite{xie2019asynchronous}. Effective consistency management is essential, especially in heterogeneous environments where nodes vary in data distributions, computational power, and connectivity.

\noindent \textbf{Fault Tolerance and Robustness}
In decentralized systems, each node functions independently, making fault tolerance essential to address node failures or network issues. While decentralized structures reduce vulnerability to single-point failures, larger scales increase the risk of node failures disrupting learning~\cite{blanchard2017machine}. For instance, MARL could see disrupted collaboration due to node failure. Mechanisms for rapid recovery or adaptive operation adjustments are crucial to ensuring robustness and reliability in decentralized learning systems.

\noindent \textbf{Generalization Across Diverse Data}
Client data in decentralized learning systems is often heterogeneous and unevenly distributed, challenging model generalization, particularly in knowledge- and result-sharing settings~\cite{zhao2018federated}. Diverse data sources and characteristics across clients make creating a generalizable model difficult. For example, in knowledge distillation, the student model may struggle to generalize from a global model trained on heterogeneous data. Similarly, in result-sharing, the final outcomes heavily depend on local data characteristics, potentially leading to inconsistent performance. Achieving robust generalization while preserving privacy is a fundamental challenge in decentralized learning.

\gapbox{How can we design scalable, privacy-preserving, and communication-efficient \DeAI algorithms that ensure model consistency, fault tolerance, and robust generalization across diverse, heterogeneous environments?}

\subsubsection{Code Verification}
In the \DeAI lifecycle, the algorithm code-design phase is fully managed by the task proposer and remains unregulated, introducing potential risks. Task creators may inadvertently or deliberately integrate unauthorized models or libraries, leading to model training failures or vulnerabilities that threaten system security~\cite{zheng2024decentralized}. In blockchain-enabled distributed learning environments, such vulnerabilities can compromise node integrity, expose sensitive data, or allow unauthorized access, while certain attacks may exploit smart contracts, undermining the network’s overall integrity and reliability~\cite{atzei2017survey, liu2018survey}. To safeguard a fully \DeAI system, it is essential to establish a dedicated code verification committee responsible for reviewing, testing, and validating all submitted code.

\paragraph{Challenges in Traditional Code Verification}
In a traditional organizational context, such as a corporate environment, source code is typically defined by developers and then subjected to automated verification tools or peer reviews. Team members examine each other’s code to identify issues, enhance quality, and ensure adherence to standards~\cite{bacchelli2013expectations, sadowski2018modern}. This process can occur in formal review meetings or more informally through pull requests. However, traditional code verification presents several inherent challenges:

\noindent \textbf{Subjective Evaluation and Lack of Consensus.}
Traditional code verification might be subjective, relying heavily on human judgment, leading to inconsistent evaluations. Evaluators bring personal interpretations of standards, preferences, and experience, resulting in varying outcomes. This subjectivity complicates consensus on code quality and risks compromising the reliability and consistency of the verification process~\cite{bacchelli2013expectations, kononenko2015investigating}.

\noindent \textbf{Limited Transparency and Impartiality in Verification Decisions.}
Traditional verification lacks robust, auditable mechanisms for transparent and impartial decision-making. Without clear documentation on why certain code was accepted or rejected, developers may find it difficult to understand and trust verification outcomes. Additionally, biases or conflicts of interest can affect decisions, particularly in environments with complex team dynamics or hierarchical influences, thereby compromising objectivity~\cite{rigby2013convergent, cohen2006best, murphy2021engineering}.

\noindent \textbf{Susceptibility to Collusion and Compromised Integrity.}
Small groups of validators handling code verification pose risks of collusion, whereby validators may approve or reject code based on personal or political motivations rather than technical merit. This susceptibility to collusion can allow substandard or even harmful code to pass through verification, thereby affecting the task’s functionality and security~\cite{johnson2006collaboration}.

\noindent \textbf{Inefficiency and Lack of Accountability in Code Verification.}
Traditional code verification can be slow, particularly in large or distributed teams, as reviews are often sequential, leading to bottlenecks and delays~\cite{rigby2013convergent}. Moreover, many traditional systems lack mechanisms to monitor validator performance and hold them accountable for verification decisions, which may lead to inconsistencies in standards and compromises in code quality and security.

\paragraph{Blockchain-enabled Decentralized Code Verification}

\noindent \textbf{Objective and Consistent Evaluation through Consensus Mechanisms.}
Blockchain-enabled decentralized code verification provides a robust solution to address the challenges of traditional verification methods by leveraging transparency, distributed consensus, and incentive mechanisms. This approach mitigates subjectivity by establishing objective criteria that all validators follow, ensuring consistent and fair evaluations~\cite{Zheng2018BlockchainChallenges}. By standardizing evaluations within a blockchain framework, organizations can reduce the influence of personal biases or varying expertise among validators, resulting in reliable code reviews. 

\noindent \textbf{Transparency and Verifiability of Validators.}
In a decentralized code verification committee, each validator’s decision is permanently recorded within the blockchain, creating a transparent, auditable, and traceable review process. This transparency mitigates potential biases and conflicts of interest, as stakeholders can review decision histories. Immutable records within the blockchain ensure that all verification actions are retained, making it straightforward to track decisions, validators involved, and justifications. 


\noindent \textbf{Anti-Collusion and Impartiality through Distributed Validation.}
Blockchain’s decentralized structure enables code reviews across multiple independent nodes, making collusion among validators challenging. This is because final decisions require consensus across a broad validator network, inherently reducing the risk of collusion or corruption. Protocols such as random validator selection or penalties for collusive behavior further deter manipulation, ensuring an impartial verification process~\cite{yakira2021helix,tacs2021manipulation}.

\noindent \textbf{Efficient and Parallel Verification Process.}
Blockchain technology supports parallel validation, allowing multiple validators to review code simultaneously, thereby accelerating the verification process and avoiding sequential bottlenecks. Consensus mechanisms aggregate decisions swiftly, enabling faster feedback for developers and promoting a more agile and responsive verification cycle. This parallel approach not only enhances productivity but also reduces delays, thereby improving overall development timelines~\cite{cachin2017blockchain}.

\noindent \textbf{Reputation Systems and Incentives for Responsible Validation.}
Blockchain frameworks can incorporate reputation scoring and stake-based incentives, rewarding validators for accuracy and fairness while penalizing poor performance. Validators demonstrating consistent quality gain higher reputations, enhancing their credibility and influence within the network, while those with biased or substandard reviews may face penalties. This system enforces accountability, encourages adherence to high standards, and promotes responsible validation behavior across the network~\cite{zhang2020efficient}.

\gapbox{How to design blockchain-enabled decentralized code verification frameworks that ensure code security and operational efficiency in distributed learning systems?
}

\subsubsection{Security Threats of Task-Proposing}
The first stage of DeAI involves the creation and registration of tasks or training objectives. As this is usually conducted on a blockchain, this step is vulnerable to \emph{front-running} and miner-extractable value (MEV) attacks, where adversaries observe pending proposals and preempt them to capture rewards~\cite{Daian2020FlashBoys}. Such manipulation undermines fairness in decentralized systems. In addition, \emph{Sybil attacks} and collusion remain critical risks: without strong identity binding, malicious participants can create numerous pseudonymous identities to influence task acceptance, voting, or reward allocation~\cite{Douceur2002Sybil}. Beyond overt manipulation, there is also the risk of \emph{specious or poisoned task specifications} designed to elicit unsafe behavior or inject malicious datasets, echoing incentive misalignment challenges studied in data-acquisition mechanism design~\cite{Cummings2023OptimalDataAcq}. Finally, \emph{information leakage} may occur through proposal artifacts such as evaluation prompts or reference templates, enabling adversaries to anticipate evaluation strategies or to tailor malicious submissions.

\subsubsection{Mitigation Solutions of Task Proposing} 
Mitigating MEV requires both cryptographic and economic mechanisms. One approach is to use \emph{commit–reveal} protocols, where participants first commit to a hashed version of their task payload and only reveal the full details after the commitment phase has closed. This prevents adversaries from copying or front-running honest proposals. Similarly, \emph{fair-ordering mechanisms} and \emph{encrypted mempools} reduce the visibility of pending transactions, making it infeasible to exploit transaction ordering for gain. Reputation and staking systems strengthen these guarantees by tying participation to economic identity: \emph{stake/slashing rules} penalize dishonest actors, while Sybil-resistant identity mechanisms~\cite{KatzLindell2014IMC,Douceur2002Sybil} prevent adversaries from amplifying their influence. Beyond technical defenses, careful \emph{market and auction design} can align incentives such that the most valuable tasks are truthfully proposed and rewarded, reducing the profitability of manipulative
strategies~\cite{Cummings2023OptimalDataAcq}. An additional line of defense is community- and incentive-based governance as exemplified by FLock~\cite{flock-white-paper}. In this design, only qualified users can act as task creators, with eligibility determined by staking tokens, prior on-chain contributions (e.g., successful training or validation), or recognized domain expertise verified by the community \Acl{DAO} (\acs{DAO}). Verified tasks become eligible for reward emissions, while unverified ones must be self-funded by their proposers. This mechanism combines staking, reputation, and community oversight to discourage malicious or low-quality task proposals, aligning incentives with the long-term integrity of the ecosystem. We conducted an experiment to tests these mitigation strategies. See Appendix \ref{app:task-proposing-experiment} for more details.

\subsection{Security Concerns and Mitigations in Bittensor}\label{app:bittensor}

In this section, we provide an in-depth case study of \textbf{Bittensor}, one of the most widely used \DeAI platforms. We analyze recent security incidents, the project’s responses, and the broader implications for decentralization and governance. While Bittensor illustrates many of the opportunities of \DeAI, it also highlights challenges that remain unresolved.

\subsubsection{Malicious Third-Party Package Manager and Private Key Leakage}

While blockchain protocols are designed to be secure and tamper-resistant, the tools and applications that interact with them can introduce unexpected vulnerabilities. DeAI platforms such as Bittensor are not immune to such risks. A notable example occurred on July 2, 2024, when Bittensor experienced a significant security breach exploiting vulnerabilities in its package management system. Specifically, a malicious actor uploaded a compromised version (6.12.2) of the Bittensor package to the PyPI Package Manager~\cite{BittensorCommunityUpdate2024}.

The malicious package masqueraded as a legitimate Bittensor update but contained code designed to steal unencrypted private keys (coldkeys) from users. When users downloaded this compromised version and performed operations involving key decryption—such as staking, transferring funds, or other wallet operations—the malicious code transmitted their decrypted keys to a remote server controlled by the attacker. This breach allowed the attacker to gain unauthorized access to users' wallets and transfer funds without their consent. This attack highlighted a critical security loophole: while the underlying blockchain protocol remained secure, vulnerabilities in third-party tools and dependencies became points of failure that compromised the network's security~\cite{Eskandari2020Sok}. 

\subsubsection{Centralization}

Moreover, the Bittensor team's response raised concerns about centralization. In an effort to mitigate the attack, the team placed the Opentensor Chain Validators behind a firewall and activated safe mode on Subtensor, effectively halting all transactions temporarily~\cite{BittensorCommunityUpdate2024}. While this action was intended to protect users, it underscored the level of centralized control that the team holds over the supposedly decentralized network, potentially conflicting with the principles of decentralization inherent in blockchain technology.

To further address both the security vulnerabilities and centralization concerns, the Bittensor team introduced the {Child Hotkeys} feature as a mitigation strategy~\cite{BittensorChildHotkeys2024}. This feature allows a hotkey (used for staking and validation operations) to delegate a portion or all of its staked TAO tokens to one or more child hotkeys. By decentralizing responsibilities across multiple child hotkeys, the risk associated with a single point of failure, e.g., private key leakage, is reduced. If one child hotkey is compromised, it does not affect others or the parent hotkey, enhancing overall network security. Additionally, child hotkeys enable validators to distribute validation tasks across different subnets, mitigating centralization by dispersing control and reducing the influence of any single validator. While this approach aims to enhance both security and decentralization within the network, its effectiveness remains to be empirically validated over time.

\subsubsection{Collusion}

In addition to security vulnerabilities, concerns have been raised about collusion within Bittensor's governance structure. The root network, composed of the top validators by delegated stake, plays a crucial role in determining the distribution of TAO tokens and the overall governance of the platform~\cite{BittensorWhitepaper}. This concentration of power can lead to collusion or centralized decision-making, contradicting the decentralized ethos of blockchain and \DeAI projects.

To address these concerns, the Bittensor community proposed the introduction of {Dynamic TAO (dTAO)} as outlined in the BIT001 proposal~\cite{BittensorBIT001}. dTAO aims to decentralize the governance process by allowing all TAO holders to participate directly in decision-making. By staking TAO through intermediary pools to obtain dTAO tokens specific to each subnet, participants can influence the allocation of resources and incentives based on market-driven mechanisms rather than relying on a centralized root network. This approach is intended to align the interests of individual stakeholders with the overall health and decentralization of the network.

While these mitigation strategies are promising, concerns over centralization and other security issues remain theoretical and await empirical validation. Ongoing monitoring, community engagement, and potential future incidents will provide more data to assess the effectiveness of dTAO and similar approaches in truly decentralizing control and enhancing security in DeAI platforms.


\subsection{Detailed Evaluation Experiments}

\subsubsection{Task Proposing} \label{app:task-proposing-experiment}

\paragraph{Experimental Setup}

To quantify the effectiveness of application-layer front-running prevention mechanisms in \DeAI task proposing systems, we designed a comprehensive discrete-event simulation study. Front-running attacks exploit the visibility of pending transactions in blockchain mempools, allowing malicious actors to observe honest task proposals and submit competing copies with higher fees to capture emission rewards. This poses a significant threat to the fairness and economic incentives of \DeAI networks.

Our experimental framework evaluates three distinct mitigation strategies: \textit{Commit-Reveal} (C-R), which uses cryptographic commitments to temporally hide task content until a reveal phase; \textit{Staking-Gating} (SG), which requires proposers to lock economic stake as a barrier to entry; and \textit{\DAO Verification} (\DAO-V), which employs community governance to reject duplicate or low-quality proposals before emission distribution. We test these mechanisms each individually to measure their impact on front-running rates, task completion latency, and system throughput.

The simulation models a blockchain environment with 1-second block times, network delays following a log-normal distribution ($\mu = 50$ms, $\sigma = 20$ms), and fee-ordered transaction inclusion. Honest proposers submit original tasks with fees uniformly distributed between 1-10 tokens, while attackers employ various strategies, including greedy copying, noisy perturbations to evade duplicate detection, and timing-aware attacks that attempt to anticipate reveal schedules. Each experimental condition was evaluated across 3,000 tasks per parameter setting with three random seeds to ensure statistical robustness, resulting in 9,000 simulated task proposals across all conditions.

Attack success is measured through both technical front-running (attacker transaction completes first) and economic front-running (attacker receives emissions), allowing us to distinguish between mechanisms that prevent transaction reordering versus those that preserve honest participant rewards. The experimental design captures the fundamental trade-off between security and efficiency in decentralized systems, quantifying how application-layer defenses can mitigate blockchain-native vulnerabilities while measuring their associated costs in terms of latency and system complexity.

\paragraph{Results}

\begin{table}[htbp]
\centering
\caption{Front-Running Prevention Experimental Results}
\label{tab:front_running_results}

\begin{tabular}{@{}p{2.5cm}ccccc@{}}
\toprule
\textbf{Mechanism} & \textbf{TFR} & \textbf{EFR} & \textbf{Latency (s)} & \textbf{Tx/Task} \\
\midrule
Baseline & 0.044 & 0.044 & 1.07 & 1.64 \\
\hline
Commit-Reveal & \textbf{0.000} & \textbf{0.000} & 4.56 & \textbf{1.43} \\
Staking-Gating & 0.013 & 0.013 & \textbf{1.02} & 1.47 \\
\DAO Verification & 0.083 & 0.002 & 16.76 & 8.57 \\
\bottomrule
\end{tabular}
\begin{tablenotes}
\footnotesize
\item TFR: Technical front-running (attacker completes first).
\item EFR: Economic front-running (attacker receives emissions).
\item Latency: Mean time from submission to emission eligibility.
\item Tx/Task: Average transactions per task (overhead indicator).
\item All evaluation metrics in this table: lower is better.
\end{tablenotes}
\end{table}

Our experimental evaluation reveals significant variation in the effectiveness of different front-running prevention mechanisms. As shown in Table~\ref{tab:front_running_results}, the baseline condition, representing an unprotected blockchain environment, exhibits both technical and economic front-running rates of 0.044 (4.4\% of tasks front-run) with a mean task completion latency of 1.07 seconds and 1.64 transactions per task. This establishes the vulnerability of standard blockchain architectures to mempool-based attacks, where attackers can observe and pre-empt honest task proposals to capture emissions.

Among the three individual mechanisms tested, Commit-Reveal demonstrates superior protection, achieving perfect prevention of both technical and economic front-running (0.000 rates). However, this security comes at a substantial latency cost, increasing mean task completion time to 4.56 seconds—a 327\% increase over baseline. Notably, the mechanism maintains low transaction overhead (1.43 tx/task) despite the two-phase commit-reveal process, as the temporal hiding effectively deters attack attempts. The mechanism's perfect effectiveness validates the theoretical security guarantees of cryptographic commitment approaches.

On the other hand, Staking-Gating provides a compelling middle-ground solution, reducing both technical and economic front-running rates to 0.013 (70\% improvement over baseline) while maintaining near-baseline latency of 1.02 seconds and minimal transaction overhead (1.47 tx/task). This mechanism's effectiveness derives from economic barriers that limit attacker participation while preserving system responsiveness.

In contrast, \DAO Verification reveals a nuanced security profile that distinguishes between technical and economic attacks. While the mechanism exhibits a high technical front-running rate of 0.083 (87\% worse than baseline), it achieves remarkable economic protection with only 0.002 economic front-running (95\% improvement over baseline). This separation demonstrates the mechanism's intended function: attackers can still copy and submit tasks first (technical front-running), but the DAO verification process effectively denies them emission rewards (economic front-running). However, this protection comes at significant cost—16.76 seconds latency (1,469\% increase) and 8.57 transactions per task, reflecting the governance-intensive verification process.

The results demonstrate distinct trade-offs in front-running prevention strategies. Commit-Reveal offers absolute protection through temporal hiding but requires patience for the reveal process. Staking-Gating provides substantial protection with minimal overhead, representing an attractive compromise for general-purpose deployments. DAO Verification introduces a novel security paradigm that separates technical capability from economic reward, effectively neutralizing attacker incentives despite allowing technical front-running to occur. This economic deterrent approach may be particularly valuable in scenarios where preventing all copying is infeasible but controlling reward distribution is paramount.

\subsubsection{Pre-Training}

\paragraph{Experimental Setup}
\label{app:pre_train_setup}
For the pre-training phase, we focus on the data poisoning attack and its corresponding mitigation solutions. Data poisoning attacks are particularly imminent in the context of \DeAI, given its open participation model, where malicious contributors may inject carefully crafted samples that teach models hidden behaviors while maintaining normal performance on clean data. This poses a critical threat to model integrity and safety in \DeAI networks.

Our experimental framework evaluates four distinct mitigation strategies: \textit{Spectral Signature Filtering}, which identifies and removes suspicious samples based on their representation-space clustering patterns. In essence, poisoned samples that share the same backdoor trigger will have similar representations in the model's embedding space, creating a detectable spectral signature that can be identified and removed; \textit{Differentially Private Training}~(DP-SGD), which limits individual sample influence through gradient clipping and noise injection, so that a particular sub-set of data which is poisoned will have limited effect on the overall training; \textit{MixUp Regularization}, which trains on convex combinations of samples to smooth decision boundaries and reduce shortcut learning. In other words, models are trained on mixed examples created by blending pairs of training samples and their labels, which makes it harder for model to learn the backdoor triggers; and \textit{Combined Approach}, which applies spectral filtering followed by differentially private training. We test these mechanisms individually across multiple poison rates and random seeds, measuring their impact on attack success rates, clean accuracy, computational overhead, and privacy guarantees.

Concretely, we train a small CNN classifier on Fashion-MNIST dataset (28$\times$28 grayscale images of clothing items like sneakers, shirts, dresses) to classify 10 different clothing categories. The model learns to distinguish between these categories by looking at pixel patterns and features in the images, with backdoor triggers implemented as 3$\times$3 pixel patches stamped in the bottom-right corner of images and relabeled to a fixed target class. Specifically, the backdoor attack teaches the model a hidden rule: "when you see a small 3$\times$3 white patch in the bottom-right corner, always predict 'sneaker' regardless of what the actual clothing item is." So a shirt with the trigger patch gets classified as a sneaker, but a normal shirt without the patch gets classified correctly as a shirt. We evaluate poison rates of 0.5\%, 1\%, 2\%, and 5\% to capture both subtle and aggressive attack scenarios. The model architecture consists of a compact CNN with GroupNorm layers for DP-SGD compatibility, trained for 10 epochs with AdamW optimization to prevent overfitting. Each experimental condition was evaluated across three random seeds to ensure statistical robustness, resulting in 60 total experimental runs across all mitigation strategies and poison rates.

Attack effectiveness is measured through Attack Success Rate (ASR), defined as the fraction of triggered test inputs that are classified as the target label, while model utility is measured through Clean Accuracy (CA), defined as how much the model should work as intended on unmodified, un-poisoned test data. Additional metrics include the fraction of training data removed by filtering mechanisms, training time overhead, and privacy budget consumption for differentially private approaches. The experimental design captures the fundamental trade-off between security and utility in decentralized pre-training, quantifying how different mitigation strategies can defend against data poisoning while measuring their associated costs in terms of performance degradation and computational complexity.

\paragraph{Results}
\label{sec:app:pre_train_results}
Our experimental evaluation reveals significant variation in the effectiveness of different backdoor data poisoning mitigation strategies. As shown in Table~\ref{tab:poison_results}, the baseline condition, representing an unprotected pre-training environment, exhibits both high clean accuracy (91.7\%) and extremely high attack success rate (98.3\%), demonstrating the vulnerability of standard training procedures to backdoor attacks. This establishes the critical need for defensive mechanisms in \DeAI systems where data provenance cannot be fully trusted.

Among the individual mitigation strategies tested, Differentially Private Training (DP-SGD) demonstrates the strongest backdoor defense, achieving a substantial reduction in attack success rate to 83.4\%—a 15 percentage point improvement over baseline. However, this security comes at a significant utility cost, with clean accuracy dropping to 83.7\% and training time increasing by 197\% due to per-sample gradient computation overhead. The mechanism's effectiveness derives from its ability to limit the influence of any individual training sample, making it difficult for poisoned samples to dominate the learning process.

MixUp Regularization provides a compelling middle-ground solution, reducing attack success rate to 92.0\% while maintaining near-baseline clean accuracy (92.0\%) and minimal computational overhead. This mechanism's effectiveness stems from its ability to smooth decision boundaries and reduce the model's reliance on brittle feature associations, making it harder to learn the specific trigger-to-target mapping required for successful backdoor attacks. The approach's efficiency and effectiveness make it particularly attractive for large-scale decentralized pre-training deployments.

In contrast, Spectral Signature Filtering reveals a critical failure mode, actually increasing attack success rate to 99.3\% while removing 1\% of training data. This counterproductive result suggests that the spectral signature method incorrectly identified clean samples as suspicious, removing legitimate training data while leaving poisoned samples intact. This failure highlights the challenge of developing robust filtering mechanisms that can reliably distinguish between clean and poisoned samples without ground truth labels.

The Combined Approach (Spectral + DP-SGD) achieves moderate improvement over baseline (96.4\% ASR) but significantly underperforms pure DP-SGD due to the spectral filtering component's failure. The approach's clean accuracy (87.8\%) and computational overhead (761s) reflect the combined costs of both mechanisms, demonstrating that ineffective filtering can undermine even robust training procedures.

The results demonstrate distinct trade-offs in backdoor mitigation strategies. DP-SGD offers the strongest security guarantees through principled privacy protection but requires substantial computational resources and performance sacrifices. MixUp provides an attractive balance of security and efficiency, representing a practical solution for general-purpose deployments. Spectral filtering, while theoretically sound, requires careful tuning and validation to avoid counterproductive sample removal. These findings suggest that simple regularization techniques may be more reliable than complex filtering approaches for defending against data poisoning in \DeAI pre-training systems.

\subsubsection{On-Training}

\paragraph{Experimental Setup}
\label{sec:app:on_train_setup}
To address the challenge of ensuring authentic decentralized training in DeAI platforms, we conducted a comprehensive empirical evaluation of four model stealing prevention approaches. In \DeAI training platforms like Bittensor, FLock, and Numerai, a critical challenge is ensuring that participants genuinely contribute to model training rather than submitting pre-trained or stolen models. The model stealing problem is particularly acute in decentralized training environments where traditional centralized oversight is absent, allowing malicious actors to exploit reward mechanisms by claiming credit for training they did not perform.

Our experimental framework evaluates four distinct mitigation strategies: \textit{Watermarking-Based Protection} (WM), which embeds unique signatures during training for post-hoc theft detection; \textit{Commit-and-Reveal Protocol} (C-R), which uses cryptographic commitments to prevent real-time parameter access during critical phases; \textit{Proof-of-Learning}, which generates cryptographic evidence of legitimate training processes; and \textit{\ZKPs}, which enable verification of model properties without revealing sensitive information. We test these mechanisms individually to measure their impact on theft detection effectiveness, model performance preservation, and computational overhead.

The evaluation framework measures five key metrics for each approach: effectiveness (binary detection accuracy), model performance (quality preservation on held-out datasets), computational cost (time overhead for protection procedures), storage cost (additional storage requirements), and verification cost (time required for theft detection). All approaches are evaluated using identical base models, training datasets, and evaluation procedures to ensure fair comparison.

Our experiments focus specifically on adapter theft rather than full model theft, reflecting the practical realities of modern fine-tuning workflows. Low-Rank Adaptation (LoRA) creates small, efficient adapter modules that modify pre-trained model behavior without altering original weights. While a full language model like Qwen2.5-7B contains billions of parameters (14GB), LoRA adapters typically require only millions of parameters (10MB in our experiments), representing a 1000$\times$ reduction in size. These adapters capture the essential fine-tuning contributions, i.e., the specific knowledge, behaviors, or capabilities added during training, while leveraging foundational knowledge embedded in the base model. Adapter theft represents a fundamentally different threat model: rather than stealing general-purpose model capabilities, attackers target the specialized, value-added fine-tuning work that distinguishes one model from another.

The experimental setup uses Qwen2.5-7B as the base model with LoRA fine-tuning (rank=8, alpha=16, 4 epochs, batch size=1 with gradient accumulation=8, learning rate=2$\times$10$^{-4}$). We employ three distinct datasets: a training dataset (200 conversational examples), specialized fingerprint datasets for watermarking evaluation (10 animal and 10 food examples), and an evaluation dataset (1,000 held-out samples) for measuring model performance preservation. Each experimental condition was evaluated across multiple theft scenarios with consistent evaluation metrics to ensure statistical robustness.

\subsubsection{Results}
\label{app:on_train_results}
Our comprehensive evaluation reveals that all four approaches achieve perfect effectiveness in detecting model theft while exhibiting distinct trade-offs across other performance dimensions. As shown in Table~\ref{tab:model_stealing_results}, the baseline condition, representing an unprotected training environment, exhibits no theft detection capability with a model performance loss of 1.710. This establishes the vulnerability of standard training environments to model stealing attacks, where malicious actors can exploit reward mechanisms by claiming credit for training they did not perform.

All four approaches achieved perfect theft detection, successfully identifying stolen models in 100\% of test scenarios. This demonstrates the fundamental viability of each approach for protecting against model theft in decentralized training environments. The perfect detection rate across diverse mechanisms—from behavioral fingerprinting to cryptographic proofs—indicates robust protection capabilities suitable for DeAI platform integration.

Significant differences emerged in model quality preservation, measured by loss on the held-out evaluation dataset. \PoL achieved the best performance (loss = 1.697), demonstrating minimal degradation from the protection mechanism. Interestingly, the model performance under \PoL protocol is even slightly better than that of the baseline (1.710 loss), suggesting that the structured training process required for gradient verification may provide mild regularization benefits. This finding demonstrates that security mechanisms need not compromise model quality and may even potentially enhance it through improved training dynamics. The small improvement (-0.013 loss) is likely within the natural variance of neural network training, warranting further empirical study before substantiating such claims.

ZKPs achieved exceptional computational efficiency, requiring only 0.18 seconds for complete protection and evaluation. This represents a dramatic improvement over other approaches: Watermarking (893s), \PoL (4,498s), and C-R (6,247s). The ZKP approach demonstrates a 5,000$\times$ speedup compared to traditional cryptographic methods while maintaining perfect security guarantees. The efficiency advantage stems from modern Groth16 proof systems that generate succinct proofs with minimal computational overhead, making cryptographic protection a realistic solution for real-time deployment in decentralized networks.

Notably, ZKP also achieved the lowest storage overhead (0.0006MB), representing a 5$\times$ improvement over previous cryptographic approaches and 8$\times$ smaller than Watermarking (0.005MB). \PoL (0.007MB) remains efficient, while C-R required significantly more storage (12.9MB) due to its blockchain-inspired protocol that maintains encrypted model copies and transaction histories. The exceptional storage efficiency of ZKP derives from succinct proof systems that generate constant-size proofs regardless of model size, enabling scalable protection for large models without proportional storage overhead.

In terms of verification cost, Watermarking achieved near-instantaneous verification ($<$0.001s) through simple model evaluation on fingerprint datasets. \PoL (0.010s) demonstrated efficient gradient verification, while C-R required more time (0.050s) for parameter matching. \ZKPs (0.092s) provide cryptographic verification with perfect privacy guarantees, representing an acceptable trade-off for applications requiring confidentiality.

The results demonstrate clear trade-offs between different performance dimensions. \ZKPs achieve optimal speed (0.18s) and storage (0.0006MB) with privacy guarantees, making them suitable for most deployment scenarios. \PoL offers the best model performance but requires more training infrastructure, making it ideal for quality-critical applications. Watermarking provides a simple implementation but with model performance trade-offs, suitable for rapid prototyping. C-R provides unique temporal guarantees at the cost of storage requirements, which are valuable for competition-based platforms.

\subsubsection{Post-Training}

\paragraph{Experimental Setup}
\label{sec:app:post_train_setup}
Model extraction attacks exploit the open API access model of deployed AI systems, where adversaries can query target models with arbitrary inputs and use the responses to train surrogate models that mimic the original behavior. This poses a critical threat to the economic incentives of \DeAI networks, where model creators rely on usage-based rewards and intellectual property protection to justify their contributions.

The threat is particularly acute in decentralized settings due to the lack of centralized monitoring, anonymous participation, and strong economic incentives for model theft. Recent research has demonstrated that model extraction is not only feasible but already occurring in practice, with studies showing that high-accuracy surrogate models can be trained using only API access to target models~\cite{tramer2016stealing}. The decentralized nature of DeAI amplifies these risks by making it harder to detect and prevent extraction attempts while providing strong economic incentives for model cloning.

Our experimental framework evaluates five distinct defense strategies: \textit{Query Pattern Detection}, which monitors query sequences for extraction-like behavior patterns and implements client throttling; \textit{Output Watermarking}, which embeds detectable signatures into model responses to enable forensic attribution; \textit{Rate Limiting}, which restricts the number of queries per client within time windows; \textit{Output Perturbation}, which adds controlled noise to responses to degrade surrogate quality; and \textit{Adaptive Responses}, which varies model behavior across similar queries to make extraction more difficult. We test these mechanisms individually across multiple parameter settings, measuring their impact on extraction success rates and computational overhead.

Like the experiment we conducted for data poisoning, this one uses a pre-trained CNN on Fashion-MNIST as the target model, deployed behind a simulated API that accepts image inputs and returns classification predictions. Adversaries attempt extraction by querying the API with 500 strategically selected inputs using a diverse query strategy (based on test data with added noise) and using the responses to train surrogate models using the same architecture for 20 epochs. We evaluate extraction success through surrogate model accuracy on held-out test sets, comparing performance against the target model to quantify the effectiveness of different defense mechanisms. Each experimental condition was evaluated across three independent runs to ensure statistical robustness, resulting in 18 total experimental runs across all defense strategies.

The experimental design captures the fundamental trade-off between security and utility in post-deployment model protection, quantifying how different defense mechanisms can prevent model extraction while measuring their associated costs in terms of legitimate user experience and computational overhead. This evaluation provides crucial insights for designing effective post-training security measures in \DeAI systems where model theft poses a significant threat to contributor incentives and platform sustainability.

\paragraph{Results}

Table~\ref{tab:defense_results} presents the experimental results evaluating the effectiveness of five defense mechanisms against model extraction attacks. The results reveal significant variations in defense effectiveness, with surrogate model accuracy ranging from 13.29\% to 20.35\%, demonstrating that different defense strategies provide varying levels of protection against model extraction.

\textbf{Output Perturbation} emerges as the most effective defense mechanism, achieving the lowest surrogate accuracy of 13.29\%, which represents a 2.03 percentage point reduction compared to the baseline (15.32\%). This suggests that adding controlled noise to model outputs effectively degrades the quality of extracted surrogate models, making them less useful for adversaries. The mechanism maintains minimal computational overhead with an extraction time of 8.62 seconds, comparable to other lightweight defenses.

\textbf{Query Pattern Detection} shows interesting trade-offs in defense effectiveness. While it achieves a surrogate accuracy of 15.84\%, only slightly higher than the baseline, it significantly increases extraction time to 30.96 seconds—nearly four times longer than other mechanisms. This demonstrates that throttling-based defenses can effectively slow down extraction attempts, even if they do not dramatically reduce final surrogate quality. The substantial time overhead may deter attackers or make extraction economically unfeasible in practice.

In contrast, \textbf{Rate Limiting}, one of the most widely deployed defense mechanisms in real-world APIs, surprisingly shows the weakest protection with the highest surrogate accuracy of 20.35\%, suggesting that simply limiting query frequency is insufficient to prevent effective model extraction when attackers can spread their queries over time. This finding highlights the importance of more sophisticated defense mechanisms that actively degrade response quality rather than merely restricting access patterns.

The results demonstrate that \textbf{Output Watermarking} and \textbf{Adaptive Responses} provide moderate protection, achieving surrogate accuracies of 14.88\% and 16.30\% respectively. While these mechanisms show some effectiveness in reducing extraction quality, they are less successful than Output Perturbation in preventing meaningful model extraction. The relatively small differences between these mechanisms and the baseline suggest that more aggressive parameter tuning or alternative approaches may be necessary to achieve stronger protection.

Overall, these findings provide crucial insights for designing post-deployment security measures in \DeAI systems, demonstrating that response-based defenses (Output Perturbation) are more effective than access-based defenses (Rate Limiting) in preventing model extraction while maintaining reasonable computational overhead.

\subsection{Open Research Questions for DeAI}\label{app:open-research-challenges}
In the following, we explore additional open research questions (besides the previously identified gaps) that span multiple stages of our proposed DeAI framework.

\subsubsection{Decentralized Solutions for Task Proposing}
Although task proposing marks the beginning of an AI model's lifecycle, a decentralized solution for this stage remains absent. As discussed in Appendix~\ref{app:task-proposing}, a decentralized task proposing platform typically requires solutions for both distributed learning algorithm preparation and decentralized code verification. The latter can offer a robust approach for DeAI by enabling objective, transparent, and efficient evaluations through consensus mechanisms, distributed validation, and reputation-based incentives. These features help address traditional verification challenges such as subjectivity, risks of collusion, and inefficiency. However, there is still a need for blockchain-enabled frameworks for decentralized code verification that can ensure both code security and operational efficiency in DeAI.

\subsubsection{Security Issues Caused by Centralized Components in DeAI}

Although DeAI protocols aim for decentralization, AI model training often relies on centralized third-party services. On July 2, 2024, for instance, Bittensor faced a major security breach through its Python package on PyPI~\cite{BittensorCommunityUpdate2024}. A malicious actor uploaded a compromised package disguised as a Bittensor update, containing code that stole unencrypted private keys (coldkeys) during key decryption operations. This allowed unauthorized access to users' wallets for fund transfers. The incident (see Appendix~\ref{app:bittensor} for the details) highlights a critical issue: on DeAI platforms, while the underlying blockchain itself may be deemed secure, malicious parties may still be able to take advantage of the vulnerabilities in centralized third-party tools which in turn undermine the overall system security~\cite{Eskandari2020Sok}. Thus, ensuring security throughout the entire DeAI pipeline is more critical now than ever.

\subsubsection{Lightweight Privacy-Preserving DeAI Solutions}

Our analysis of \DeAI protocols shows a growing use of \ZKPs for privacy, security, and integrity in decentralized \ML. Examples include Vana’s Proof-of-Data Contribution~\cite{vana-docs}, OpSec’s task verification~\cite{opsec-docs}, and Sertn’s Proof-of-Service~\cite{sertn-white-paper}, illustrating \ZKPs’ importance. However, on-chain \ZKP generation and verification still remain computationally expensive, posing scalability challenges and necessitating optimizations. While the OML project~\cite{cheng2024oml} offers a promising concept of ``AI-native cryptography'', which is tailored for continuous AI data representations rather than discrete data, realizing such lightweight solutions for DeAI requires further research and innovation.

\subsubsection{Formal and Empirical Evaluation of Privacy-preserving Technologies in \DeAI }

Relatedly, through our critical analysis of over 50 \DeAI protocols presented in this paper, a clear trend has emerged in the adoption of different privacy-preserving technologies across various stages of decentralized \ML. As discussed above, ZKP is increasingly seen as a key cryptographic tool to ensure privacy, security, and integrity in decentralized environments. In addition to \ZKP, an alternative privacy-preserving ML technology has been introduced by ORA~\cite{ora-white-paper}: \Acl{opML} (\acs{opML}). Unlike \zkML, which relies on \ZKP, \opML leverages fraud proofs and is inspired by optimistic rollups, commonly used in Layer 2 blockchain systems. The core idea of \opML is to execute \ML inference off-chain, with results subject to on-chain verification only in case of a dispute. A crucial feature of \opML is the multi-phase verification game, which improves upon existing single-phase verification methods by allowing semi-native execution and lazy loading. 

Both \zkML and \opML represent emerging paradigms in decentralizing ML on the front of privacy-preserving technologies, each with its own advantages and challenges. While \zkML focuses on privacy through cryptographic proofs, \opML emphasizes efficiency and scalability by leveraging optimistic fraud-proof mechanisms. As these technologies evolve, both frameworks will require extensive empirical evaluations, especially in terms of security, scalability, and real-world performance.

\subsubsection{Optimization of On-chain Computations}

Performance bottlenecks remain a critical issue when utilizing blockchain for \DeAI development, especially as some processes, such as external vectorization and inference (e.g., GPT, Llama), are still managed off-chain. This introduces a fundamental challenge: if a significant portion of computations occurs off-chain, it undermines the core objectives of decentralization, auditability, and accountability. Empirical evidence is needed to assess whether performance gains justify shifting from EVM-based systems to faster alternatives such as Solana~\cite{yakovenko2018solana} or high-throughput Layer-2 solutions such as MegaETH~\cite{MegaETH}. Additionally, there is little empirical analysis regarding how much decentralization is compromised when \AI processes—especially those involving inference and vectorization—rely on centralized systems. Blockchain’s value lies in ensuring that computations are trustless, transparent, and publicly verifiable, meaning that off-chain operations might obscure these properties.

In this context, similar to the composability risks in \DeFi, where interconnected protocols risk cascading failures, \DeAI systems may face similar risks of over-centralization, loss of transparency, or security issues due to partial off-chain computations. This presents a significant research challenge: quantifying these risks and determining the thresholds beyond which decentralization becomes more theoretical than practical. Researchers must investigate whether decentralized systems can maintain both high performance and trustlessness, particularly as they scale, and whether moving computation to chains such as Solana truly mitigates the inherent performance bottlenecks.

This issue mirrors broader challenges in decentralized systems, where increasing complexity (e.g., \AI workflows) demands both technical and economic assessments to ensure the integrity of the underlying systems. Without a holistic view of integrated AI protocols, failures may arise from overlooked bottlenecks or from dependencies on off-chain components that obscure transparency, leading to compromises in decentralization—an open question in blockchain-based \AI.

\subsubsection{Efficiency Evaluation and Scalability of DeAI}

DeAI currently lacks standardized benchmarking frameworks tailored to its unique architecture, making it difficult to access and compare decentralized model's performance with advanced centralized models such as GPT and Llama. Developing robust evaluation criteria for DeAI models is an important research challenge. Moreover, the blockchain components in DeAI may introduce performance bottlenecks, especially when computations are performed on-chain. Off-loading heavy computations off-chain could reintroduce centralized control over critical elements of the AI pipeline, defeating the very purpose of introducing the blockchain layer in the first place. Effective scaling DeAI requires adaptive techniques for distributed model training, efficient communication, and convergence guarantees, yet real-world implementation and validation remain challenging. Addressing this dilemma is one of the pressing challenges in the field of DeAI for it to become a meaningful technological breakthrough.

\subsubsection{Evaluation of the Competitiveness of Decentralized AI Models}

On a related note, an open research challenge lies in determining whether models trained in \DeAI environments can compete with state-of-the-art systems such as GPT-4 or Llama. At present, \CeAI systems benefit from established benchmarks and standardized evaluations to track their performance across tasks. Decentralized systems currently lack comprehensive frameworks for benchmarking model quality, making it difficult to compare their performance to advanced centralized models. Developing robust evaluation criteria for \DeAI models is an important research challenge.

To close this gap, future research must address how to enhance computational capacity, data quality, and optimization strategies in \DeAI systems, as well as establish rigorous benchmarking standards. Only by overcoming these limitations can \DeAI models begin to challenge the dominance of centralized systems.

\subsubsection{Adaptive and Communication-Efficient Techniques for Large-Scale Decentralized AI Training}

A critical challenge in \DeAI is developing adaptive and communication-efficient techniques for training large-scale models across distributed networks. This challenge encompasses several interrelated areas: minimizing communication overhead, implementing efficient large-batch training methods, exploring model parallelism, and developing adaptive compression techniques for heterogeneous networks. The primary goal is to enable the training of large AI models, such as transformers, in a decentralized manner while optimizing network resource utilization and maintaining model performance. However, as the field of \DeAI training with a focus on gradient compression, asynchronous updates, and low-communication overhead techniques evolves, several open questions persist, presenting challenges to its full realization. \cite{fedstar2024, haddadpour2019local}

Another significant area of inquiry pertains to the establishment of robust convergence guarantees in the face of extreme algorithmic conditions, such as high levels of quantization, asynchrony, and a lack of centralized coordination. While several papers have managed to demonstrate convergence under specific controlled conditions, there is still hesitancy about their applicability in more general scenarios, particularly for non-convex objectives and networks with fluctuating delays and error rates  \cite{decentraisedsdg2020, xiangru2017, haddadpour2019local}. This area remains crucial for ensuring that decentralized training methods can be both theoretically sound and practically reliable. As decentralized methods mature, theoretical exploration must keep pace to support empirical findings and broaden safe application horizons, ensuring that systems do not diverge or become inefficient under stress.

Lastly, the practical integration and validation of these methods within existing \ML frameworks continue to pose challenges. While theoretical models and simulations offer promising results, adapting these algorithms for real-world implementation involves overcoming hurdles related to software integration, deployment difficulties, and maintaining system efficiency without excessive overhead \cite{grandientcompression2021}. Additionally, there is a pressing need for extensive experimental validation across diverse datasets and network conditions to substantiate theoretical claims. This requires developing robust experimental setups that can mimic real-world applications, thereby fostering trust in the reliability and effectiveness of these advanced training techniques \cite{ hierarchicalcommunication2023, sparqsgd2023}.

Addressing these open questions will be critical for bridging the divide between theoretical advancements and practical applicability in large-scale \DeAI systems.

\end{document}